\documentclass[plain,ROB]{nowfnt} %KG
\journalbibstyle{numeric}%KG

\catcode`\@=11

\newcommand*{\@quotelevel}{}%
     \newcount\@quotelevel
     \newcommand*{\@quotereset}{}%
     \newcount\@quotereset
     \newcommand*{\@setquotesfcodes}{}%
     \let\@setquotesfcodes\relax
     \newrobustcmd*{\initoquote}{\@quotelevel\@ne}%
     \newrobustcmd*{\initiquote}{\@quotelevel\tw@}%
     \newrobustcmd*{\textooquote}{``}%
     \newrobustcmd*{\textcoquote}{''}%
     \newrobustcmd*{\textoiquote}{`\relax}% block ligs
     \newrobustcmd*{\textciquote}{'\relax}% block ligs
\newrobustcmd*{\enquote}{\@ifstar\blx@enquote@ii\blx@enquote}%
     \def\blx@enquote{%
       \ifnum\@quotelevel>\z@
         \expandafter\blx@enquote@ii
       \else
         \expandafter\blx@enquote@i
       \fi}%
     \long\def\blx@enquote@i#1{%
       \begingroup\initoquote
       \textooquote#1\textcoquote
       \endgroup}%
     \long\def\blx@enquote@ii#1{%
       \begingroup\initiquote
       \textoiquote#1\textciquote
       \endgroup}%

\catcode`\@=12

\usepackage{geometry}
\usepackage{pdflscape}
\usepackage{longtable}
\usepackage{soul}
\usepackage{color}
\usepackage{array}
\newcolumntype{L}[1]{>{\raggedright\let\newline\\\arraybackslash\hspace{0pt}}m{#1}}
\newcolumntype{C}[1]{>{\centering\let\newline\\\arraybackslash\hspace{0pt}}m{#1}}
\newcolumntype{R}[1]{>{\raggedleft\let\newline\\\arraybackslash\hspace{0pt}}m{#1}}

\newcommand{\ie}{\emph{i.e.}}
\newcommand{\eg}{\emph{e.g.}}
\newcommand{\etc}{\emph{etc}}
\newcommand{\now}{\textsc{now}}

\title{Embodiment in Socially Interactive Robots}

\maintitleauthorlist{
Eric Deng\\
University of Southern California\\
denge@usc.edu\\
\and
Bilge Mutlu\\
University of Wisconsin--Madison\\
bilge@cs.wisc.edu\\
\and
Maja J Matari\'{c}\\
University of Southern California \\
mataric@usc.edu
}

\issuesetup
{%
copyrightowner={E.~Deng, B.~Mutlu, and M.~Matari\'{c}},
volume = 7,
issue = 4,
pubyear = 2019,
isbn = 978-1-68083-546-5,
eisbn = 978-1-68083-547-2,
doi = 10.1561/2300000056,
firstpage = 251,
lastpage = 356
}

\usepackage{mwe}

\author[1]{Deng,Eric}
\author[2]{Mutlu,Bilge}
\author[3]{ Matari\'{c},Maja J.}
\affil[1]{University of Southern California; denge@usc.edu}
\affil[2]{University of Wisconsin--Madison; bilge@cs.wisc.edu}
\affil[3]{University of Southern California; mataric@usc.edu}

\articledatabox{\nowfntstandardcitation}%KG

\begin{document}

% the following settings can be set or left blank at first
%\copyrightowner{E.~Deng, B.~Mutlu, and M.~Matari\'{c}}
% \volume{1}
% \issue{3}
% \pubyear{2014}
% \copyrightyear{2013}
% \isbn{978-0521833783}
% \doi{1234567890}
% \firstpage{23}
% \lastpage{94}

\makeabstracttitle

%KG
%\frontmatter  % title page, contents, catalog information
%\maketitle
%\tableofcontents
%\mainmatter

\begin{abstract}
Physical embodiment is a required component for robots that are structurally coupled with their real-world environments.  However, most socially interactive robots do not need to physically interact with their environments in order to perform their tasks. When and why should embodied robots be used instead of simpler and cheaper virtual agents?

This paper reviews the existing work that explores {the role of physical embodiment in socially interactive robots}. This class consists of robots that are not only capable of engaging in social interaction with humans, but are using primarily their social capabilities to perform their desired functions. Socially interactive robots provide entertainment, information, and/or assistance; this last category is typically encompassed by socially assistive robotics. In all cases, such robots can achieve their primary functions without performing functional physical work.

To comprehensively evaluate the existing body of work on embodiment, we first review work from established related fields including psychology, philosophy, and sociology. We then systematically review {65} studies evaluating aspects of embodiment published from {2003 to 2017} in major peer-reviewed robotics publication venues. We examine relevant aspects of the selected studies, focusing on the embodiments compared, tasks evaluated, social roles of robots, and measurements. We introduce three taxonomies for the types of \textit{robot embodiment}, \textit{robot social roles}, and \textit{human-robot tasks}. These taxonomies are used to deconstruct the design and interaction spaces of socially interactive robots and facilitate analysis and discussion of the reviewed studies. We use this newly-defined methodology to critically discuss existing works, revealing topics within embodiment research for social interaction, assistive robotics, and service robotics, in which more extensive exploration would greatly improve the current understanding of the impact of embodiment on human perception and evaluation of human-robot interactions.

The introduced taxonomy for embodiment design is used as a starting point for outlining our characterization of the design space of robot embodiments. The presented characterization can be used to discuss how the physical embodiment of socially interactive robots relates to social capabilities and affordances. By introducing a general model of the design space, existing research findings can better advise robot designers and we discuss how these findings can inform researchers through design decisions in the development of future socially interactive robots.

\end{abstract}

\chapter{Introduction}
\label{c-intro}
As technology development and sophistication continue to progress at an ever-growing rate, automated systems are quickly becoming integrated into everyday life. These systems have assisted humans in tasks ranging from scheduling \citep{blum1997selection}, ordering food deliveries \citep{simmons1997layered}, entertaining guests \citep{breazeal2004designing}, enhancing assembly line work \citep{simmons2001first}, and coaching physical and mental health activities \citep{langen1994remote}.

A growing subset of these technologies are artificial agents, whether they be on-screen, in virtual reality (VR), or physically embodied. We are witnessing parallel and synergistic growth of the core technologies of artificial intelligence, computing, and manufacturing, all facilitating the development of interactive artificial agents. Researchers and engineers working in human-robot interaction (HRI) and {socially interactive robotics} are designing, building, testing, and deploying robots that interact with humans and perform a wide range of tasks \citep{goodrich2007human} as partners in a growing number of domains including manufacturing \citep{asfahl1992robots}, healthcare \citep{inoue2008effective,robins2006does,wada2006robot,werry2001can,nikolopoulos2011robotic}, education \citep{saerbeck2010expressive,greczek2014socially,clabaugh2015designing,takeuchi2006comparison,kanda2004interactive,gordon2015can}, and entertainment\citep{kidd2004effect,shinozaki2008construction,pereira2008icat,klamer2010adventures}.

As these robots are interacting with users through primarily non-physical means, it is critical for them to be able to engage in effective social interactions. Embodiment provides the opportunity to leverage more channels of communication, including proxemics \citep{takayama2009influences,mead2016perceptual,mead2013automated}, oculesics \citep{mutlu2012conversational,andrist2012designing,andrist2012designing}, and gestures \citep{breazeal2005effects,sidner2005explorations} to enhance communication and the perception of being more trustworthy \citep{reilly1996believable}, helpful \citep{reilly1996believable}, and engaging\citep{kidd2004effect} than disembodied agents.

Although embodiment is a defining feature of robotics, the study of embodiment and embodied behavior predates robotics and extends well beyond it; it spans many fields of study, including neuroscience \citep{edelman2004wider}, philosophy \citep{hendriks1996catching}, and social sciences \citep{gover1996embodied,kant1981universal}.

How critical is the physical embodiment of a robot in human-machine interaction? Embodiment is clearly a necessity for robots that physically interact with and manipulate objects, but most socially interactive robots do not physically interact with the environment to achieve their goals \citep{lee2004can,fong2003survey}. As a result, in such contexts, the benefits of physical embodiment over less expensive and complex virtual presence is less obvious \citep{holz2009robots}. This work explores the embodiment hypothesis in socially interactive robotics: \textbf{``the hypothesis that physical embodiment has a measurable effect on performance and perception of social interactions''} \citep{wainer2006role}.

\looseness=+1
Research in human communication and psychology has explored both physical and virtual embodied cues as tools for improving social interaction, including gaze behavior \citep{bailenson2001equilibrium}, head movements \citep{bailenson2006longitudinal}, and the persona effect \citep{moundridou2002evaluating}: the affective impact of artificial agents in social interaction. Kantian philosophy introduced the concepts of the mind-body and subject-object problems in relation to the embodied view in the mid-1700's \citep{gover1996embodied,kant1981universal}, leading to the development of the ``modern'' embodiment hypothesis outlined by \citet{ortega2010vitalidad}, \citet{heidegger1973art}, and \citet{merleau2004world} and \citep{fieser2011internet,benner1994interpretive}. Embodied cognition spans these fields, bringing together the work of \citet{brooks1990elephants} and \citet{moravec1988mind} in robotics and sensing, the modern-day philosophy of \citet{clark2008supersizing, clark2007re} and \citet{hendriks1996catching}, and research in neuroscience and biology from \citet{edelman2004wider}, \citet{longo2008embodiment}, \citet{damasio1999feeling}, and \citet{rosch1991embodied}. In human-computer interaction, non-physical interactions with artificial agents in social interactions have been studied \citep{cassell1999embodiment}, specifically exploring the design of such systems for social abilities and quality of interactions they can produce \citep{rehnmark2005robonaut,kramer2005social}. In robotics, specific dimensions of social interaction have been explored, as has the influence of the design of physical embodiment on interaction \citep{wainer2007embodiment}, engagement \citep{kidd2004effect,takayama2009influences,walters2005influence}, trust \citep{bickmore2001relational,bickmore2005establishing}, and the perception of an agent \citep{burgoon2000interactivity,jung2004effects,kidd2004effect,takayama2009influences,wainer2006role,walters2005influence}.

Previous work in robotics suggests physical embodiment can increase engagement and enjoyment in social interactions with humans \citep{bainbridge2011benefits,kidd2004effect,wainer2006role,wainer2007embodiment}. This paper presents a thorough review of existing work and analyzes existing results and approaches to embodiment to determine the current state of the embodiment hypothesis. As research continues to validate the importance of embodiment in \emph{socially interactive robots}, the implications on robot design will become more apparent, because both the theoretical and practical importance of physical embodiment for human-robot interactions translates into real-world applications through  appropriate embodiment design. In this meta-review, we study various robotic platforms, most of which were designed for research uses, and then adapted to task-specific applications within research studies. We explore these embodiments and approaches \citep{mutlu2012conversational} to collecting data toward quantifying the subjective qualities of the robot's physical embodiment. We then describe our characterization of \emph{the design space for socially interactive robots} toward informing both future designers and researchers.

\looseness=+1
The rest of this paper is organized as follows. We first discuss the definition of embodiment in relevant fields of study, review past work in related fields, and introduce terminology for the rest of the paper. We then introduce a taxonomy of robot embodiments that provides the contexts for human-robot interactions in the surveyed studies. We then discuss the current state of the embodiment hypothesis in socially interactive robots based on the existing body of work, provide suggestions of areas that need further exploration, and recommend approaches that aid in the design of more structured studies. Finally, we introduce a characterization of the design space of socially interactive robots, discuss how different components of a robot's design relate to aspects of social interaction, and present an approach to leveraging existing research to design or select robot embodiments for future work.

\chapter{What is Embodiment?}
\label{c-embodiment}
\textbf{Embodiment} is a fundamental concept studied in philosophy, psychology, neuroscience, communications, and engineering \citep{csordas1994embodiment,goodwin2000action,csordas1990embodiment}. In this chapter, we review how embodiment is treated in these fields to better understand the past, present, and future of the concept as it relates to socially interactive robotics.

\section{Embodiment in Philosophy and Ethics}
\label{s-embodphil}

\textbf{Embodied, or situated, cognition} is a concept derived from embodiment in philosophy and ethics, a well-studied area in the humanities that spans topics such as \textit{social interaction}, \textit{social influence}, and \textit{decision-making} \citep{shapiro2010embodied}. \citet{wilson2002six} and \citet{anderson2003embodied} discussed embodied cognition as an approach to examining the human experience being impacted by ``aspects of the body beyond the brain.''

In \textit{philosophy}, cognition is seen as being critically influenced by all aspects of an agent's body, and the discussion of embodiment in that context is focused on the agent's sensorimotor capabilities \citep{rosch1991embodied}. For example, \citet{wilson2011embodied} attributed an agent's ``beyond-the-brain body'' as playing a critical role in that agent's cognitive processes.

Embodiment is closely related to the agent's various expectations and limitations. All agents are in some way constrained by their embodiment; they are also highly dependent on affordances, ``the fundamental properties of a device that determine its way of use'',   which are themselves derived from embodiment \citep{gibson1982concept}. The affordances, expectations, and limitations set by an agent's embodiment are further discussed in Section 4.

The ethics of embodiment in social interaction relate to these affordances stemming from a robot's design. Interactive robots are often designed with the goals of being engaging and assistive. The robot's quality of being engaging aids interaction, but can also potentially lead to undesirable influence, unrealistic expectations, and perceived deception, disappointment, or emotional discomfort. Attachment toward the robot can develop, so that the removal of the robot may lead to grief and anxiety \citep{passman1975mothers}. Misleading embodiment design can also engender inappropriate use that can potentially lead to emotional or physical injury \citep{norman1999affordance}.

Classical works in philosophy establish the foundation for embodiment in general, including robot embodiment, and ethics further warns of negative consequences of some design choices.

\section{Embodiment in Psychology and Communication}
\label{s-embodpsychcomm} % another label
Scholars in the fields of psychology and human communications have long pondered the question of how and to what extent different media can be used to represent the real world. A significant body of literature discusses virtual reality \citep{merchant2014effectiveness,saposnik2011virtual}, perceived reality \citep{jussim1991social,jackson1985meta}, pictorial realism \citep{welch1996effects}, and other related topics. Increasingly, communications researchers are becoming interested in \textit{presence} and its relationship to embodiment \citep{biocca1997cyborg,mantovani1999real,lombard1997heart}. Effective design of an embodied robot serves to increase its \textit{social presence} and desired affordances.

Recent embodiment research in human communications fields has focused on presence in virtual reality platforms \citep{klein2003creating} and telepresence \citep{kim1997telepresence,durlach2000presence}, building on the classical works \citep{gunawardena1997social,gunawardena1995social}.
There have also been further explorations of physical embodiment in social agents \citep{jung2004effects,leyzberg2012physical,kidd2004effect}. In the next section, we discuss robotics embodiment studies whose results support the importance of social presence in both human-human and human-robot interaction.

\section{Embodiment in Robotics and Design}
\label{s-embodrobot}
Research in artificial software agents encompasses virtual agents, relatable agents, affective agents, and most recently chatbots, and has been focused on the development of communicative systems that are physically disembodied, such as text interfaces \citep{looije2010persuasive}, animations \citep{bartneck2004your}, or high-fidelity virtual characters \citep{devault2014simsensei}. The value of physical embodiment of artificial agents comes from the improvements seen in the interactions held between such agents and their human interaction partners. Two basic questions arise: (1) do physically embodied agents interact more effectively than their non-physically embodied counterparts? and (2) if so, why?

\citet{rosch1991embodied} discussed the influence that sensorimotor capabilities have on an agent's relationship with its environment--providing a richer experience for the agent and allowing the agent to exist in a richer context that bridges biology, psychology, and culture. \citet{brooks2002flesh} drew parallels from this philosophical generalization of embodiment to the field of robotics. The sensorimotor capabilities of biological beings and robots, at a high level, affect the agent in very similar ways: the sensors and effectors define limitations to the ability of the agent to sense, manipulate, and navigate its environments. Biological agents have brains, muscles, and nerves that communicate on a network that enables the system to function. When discussing the embodiment, or ``physical instantiation'',   of robots, we focus on the bodily presence of those machines. This includes the internal and external mechanical structures, embedded sensors, and motors that allow them to interact with the world around them \citep{brooks2002flesh,brooks1990elephants}. All components of embodiment are inherently tied to the agent's function, whether the agent is biological or artificial.

\begin{figure}
  \centering
    \includegraphics[width=\textwidth]{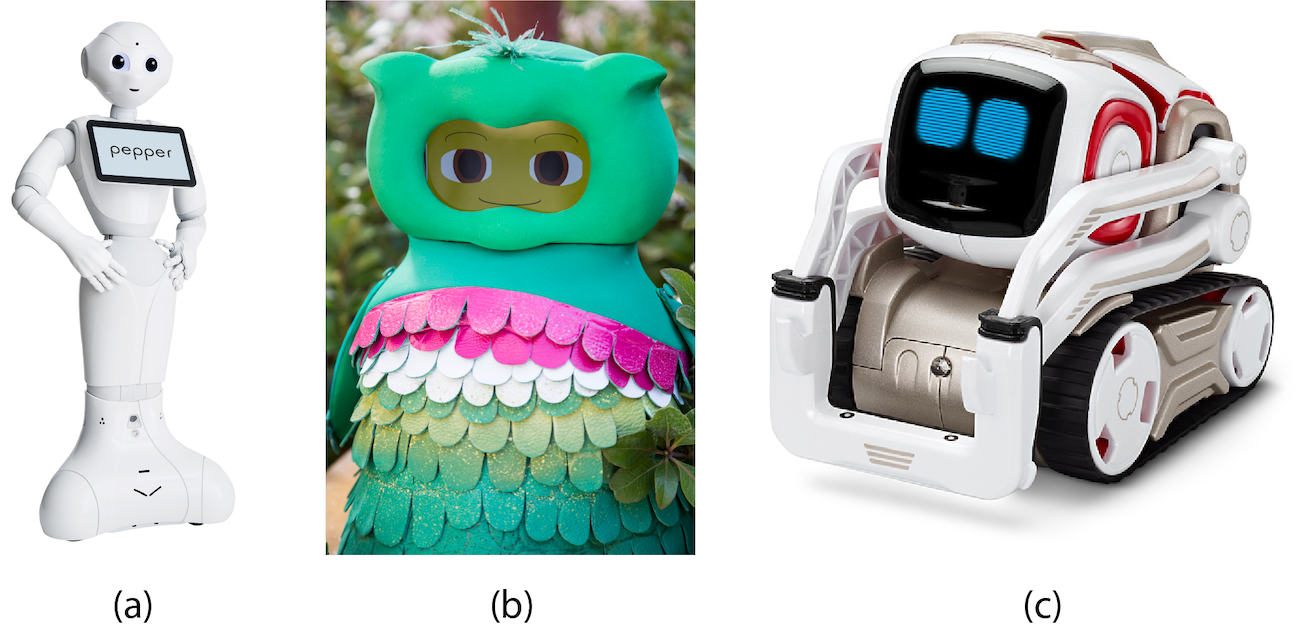}
  \caption{Embodied Socially Interactive Robot Platforms--(a) Softbank Pepper, (b) Spritebot Kiwi, (c) Anki Cozmo}
  \label{fig:example-robots}
\end{figure}

Traditionally, roboticists were largely focused on the functional properties of physical embodiment, such as locomotion \citep{furusho1986control,brooks1989robot,wang2002simultaneous}, manipulation \citep{mason1985robot,rivin1987mechanical}, and haptics \citep{lee1999review,fu1987robotics,moravec1988sensor}. Human-robot interaction (HRI) is a relatively new and rapidly growing area of robotics that focuses on interaction with people in a broad variety of settings, and fundamentally changes how value is attributed to different components in robotic systems \citep{lee2006physically,dym2005engineering,stickdorn2011service}. For instance, in HRI, the value of a gripper goes beyond its capabilities for manipulation to its role in communication: having independently-controlled fingers allows a robot hand to gesture in more complex ways and therefore opens doors to a broader realm of interactions. The value associated with socially interactive capabilities has stimulated new robot embodiments that are not capable of traditional  functions (such as Pepper, Kiwi, and Cozmo), shown in Figure \ref{fig:example-robots}).

Designing for interaction rather than physical function fundamentally changes the nature of robot design.  In Section 2, we provide a characterization of this new design space.  As a first step, we define embodiment in the context of socially interactive robots.

\subsubsection{Defining Embodiment for Interactive Agents}
\indent Embodiment in the context of artificial social agents has been a topic of discussion since the late 1900s--\citet{zlatev1997situated} explored situated embodiment, \citet{sharkey2001mechanistic} studied mechanistic and phenomenal embodiment, \citet{ziemke1999rethinking} addressed natural embodiment, and \citet{barsalou2003social} discussed the concept of social embodiment, among many others.

\citet{ziemke2003s} introduced six different ``notions'' of embodiment:
\begin{enumerate}
   \item \textbf{Structural Coupling:} The physical coupling between the agent and its environment, based on the work of \citet{maturana1991autopoiesis,maturana1987tree}. \citet{quick1999bots} provided a definition of embodiment related to structural coupling:
   \begin{quote}
   System X is embodied in an environment E if perturbatory channels exist between the two. That means, X is embodied in E if for every time t at which both X and E exist, some subset of E's possible states with respect to X have the capacity to perturb X's state, and some subset of X's possible states with respect to E have the capacity to perturb E's state.
   \end{quote}
   \item \textbf{Historical Embodiment:} The inherent relationships of any agent's embodiment with its history, especially in the context of adaptation, evolution, and growth \citep{valera1991embodied,riegler2002cognitive,ziemke1999rethinking}.
   \item \textbf{Physical Embodiment:} The physical instantiation of an agent in its environment, adapted into the concept of ``physical grounding'' \citep{brooks1990elephants} which argues that ``it is necessary to connect [intelligent systems] to the world via a set of sensors and actuators.''
   \item \textbf{Organismoid Embodiment:} The notion that cognition in an embodied artificial agent is, to some degree, dependent on its similarities to organismic counterparts.
   \item \textbf{Organismic Embodiment:} The concept that ``cognition is not only limited to physical, organism-like bodies, but in fact to organisms, i.e., living bodies'' \citep{ziemke2003s}.
   \item \textbf{Social Embodiment:} The idea that the embodiment of a socially interactive agent plays a significant role in social interactions. \citet{barsalou2003social} described social embodiment as ``states of the body, such as postures, arm movements, and facial expressions, arise during social interaction and play central roles in social information processing.'' This is the notion of embodiment more relevant to the work in this paper, as it relates most to physical embodiment of socially interactive robots.
\end{enumerate}

\citet{quick1999bots} discussed embodiment in the context of structural coupling, addressing how embodiment is presented independent of any ontological context. This notion of embodiment, inspired by the interactions of {\it Eschericha coli} ({\it E. coli}) and its environment, is most concerned with the structural or physical relationships between the agent and its surrounding environment.

This work focuses on how the physical relationship between a socially interactive robot and its surrounding environment relate to the robot's sociability and presence. We adhere to the definition of embodiment that is a combination of the concepts of ``social embodiment'' and ``situated structural coupling'' from \citet{ziemke2003s} and \citet{quick1999bots}, respectively.

In the following sections, we review the work related to embodiment in research areas outside of robotics, and discuss how they relate to the design and implementation of physical embodiment of \emph{socially interactive robots}.

\subsection{Virtual Artificial Agents}
\label{s-VH} % another label
Virtual artificial agents generally fall into one of two categories: (1) immersed virtual reality or (2) on-screen virtual characters \citep{hillis1999digital}. Virtual artificial agents and socially interactive robotics share several enabling technologies, including machine vision, speech, AI, and machine learning \citep{gratch2015exploring}. They also share related theoretical grounding, including psychology and sociology theories; \citet{persson2001understanding} presented a user-centered viewpoint of socially interactive agents, research that aims not to simulate social intelligence but to give the impression of the agent being socially intelligent. \citet{taylor2002living} explored presence, social integration, and communication in virtual worlds with secondary characters--all critical aspects of embodiment.

The research in virtual agents has shown a significant need for embodiment in virtual social interactions \citep{ruhland2015review}. Many studies supplement qualitative interviews and observer notes with quantitative data \citep{biocca2001plugging,smith2001editorial,dautenhahn2001editorial}, such as toward understanding ownership of sub-components of embodiment\citep{kilteni2012sense} and administering POMS questionnaire before and after completing an activity\citep{garau2005responses}. Virtual agents have been shown in research experiments to be engaging to a variety of populations \citep{cassell2001embodied,gratch2015exploring}, and have been developed for applications in education \citep{wagner2006real,traum2008multi}, collaboration \citep{perlin1996improv,vinayagamoorthy2004eye,richard2001inviwo,vosinakis2001simhuman}, social skill training \citep{chollet2015public}, and post-traumatic stress disorder \citep{rizzo2010development} and depression therapy \citep{devault2014simsensei}.

Work in virtual agents has repeatedly demonstrated the positive effects of embodied cues, such as gestures and expression \citep{scherer2012perception}, in maintaining user engagement in both short-term and long-term interactions \citep{kose2009effects,bickmore2005establishing,bickmore2005social}. Such embodied cues are shared aspects of virtual agents and socially interactive robots and have been shown to be transferable \citep{ono2001model}.

\subsection{Collaborative Robots}
\label{s-CoRo} % another label

Virtual agents can enable \textit{copresence}, but physical robots enable \textit{colocation}, which, in turn, can enable collaboration.  As HRI expands, physical collaboration between people and machines is a major target of research and applications, ranging from manufacturing to the service sector.

Robots were originally envisioned for performing the three ``Ds'': dirty, dull, and dangerous work \citep{murphy2000introduction}. One of the first uses of robots at scale was in manufacturing and automation. Because of the predictability and repeatability of tasks on the assembly line, robots were designed for and placed in environments not accessed by human workers, or were caged for safety. Recent research and technological advancements have enabled the development of robot systems and control algorithms for deployment in manufacturing settings where people and robots share the same environment and work together to accomplish common goals \citep{nikolaidis2013human}. This development initially focused on intuitive interfaces and communication tools for master-slave relationships between operators and robots in teleoperational situations \citep{strabala2013towards,mainprice2013human}, but such systems required trained professionals to operate them and had marginal impact on the efficiency and safety of industrial workplaces.

To allow for less trained users to effectively interact with and leverage industrial robot systems, HRI researchers have been working on various approaches to human-robot collaboration, such as using cross-training to improve task sharing between human and robot workers \cite{nikolaidis2013human}, planning shared work plans taking human ergonomics into consideration \cite{pearce2018optimizing}, and adapting robot actions to human motion, availability, adaptability, and intent \cite{huang2016anticipatory,lasota2015analyzing,huang2015adaptive,nikolaidis2017game}.

\subsection{Service and Socially Interactive Robots}
\label{s-SIR} % another label

Concurrently with collaborative robotics, the broad area of \emph{service robotics} has been growing rapidly, developing robots that can provide services in everyday life, such as vacuuming and cleaning floors \citep{jones2006robots,forlizzi2006service,mutlu2008robots}, folding laundry \citep{osawa2006clothes,maitin2010cloth}, delivering packages \citep{simmons1997layered,coltin2014online}, giving museum tours \citep{nourbakhsh1999affective},  driving autonomously \citep{levinson2011towards}, and providing aid to special needs populations in the context of socially assistive robotics \citep{feil2005defining,bemelmans2012socially,broekens2009assistive}, along with numerous other uses.

As robots move from cages and from behind closed doors into shared spaces with humans, is has become critical to integrate social capabilities and new design considerations into the embodiments of those systems.  Some considerations are related to safety, such as hiding pinch points and adding in physical compliance, and others to practical usability, such as height adjustment \citep{haddadin2009requirements,wyrobek2008towards,matthias2011safety}.

Similar to collaborative robots, socially interactive robots also need to be designed to be minimally intrusive, but their embodiments are used as tools for communication, acceptance, and engagement.  These robots primarily interact through their social capabilities in order to achieve their goals. Accordingly, they must be able to both perceive \citep{rani2004anxiety,kennedy2007spatial,bauer2008human,scherer2012perception} and generate communicative signals \citep{ono2001model,huang2014learning} that their human counterparts are able to intuitively understand, relate to, and accept.  These requirements mean a fundamental change in the way robot embodiments are designed.

\subsubsection{Social Performance and Social Presence in Embodied Robots}

The combined ability of an artificial agent to generate and understand verbal and non-verbal communication can be organized into the following seven \textit{human social} characteristics that can greatly improve a robot's social acceptance \citep{fong2003survey}:
\begin{enumerate}
   \item Express emotion
   \item Communicate with high-level dialogue
   \item Learn/recognize models of other agents
   \item Establish/maintain social relationships
   \item Use natural cues (gaze, gestures, etc.)
   \item Exhibit distinctive personality and character
   \item Learn/develop social competencies
\end{enumerate}

These components of social interaction tie into the concept that \citet{lee2006physically} referred to as \textit{social presence}, a key component in the success of social interactions.  Studies have shown that physically-embodied agents possess social presence to a greater extent than their virtual counterparts \citep{shinozawa2003robots,heerink2010assessing}.

There are a few different definitions of social presence across related research communities (HCI, communications, etc.); we adhere to the definition by \citet{bainbridge2011benefits} that defines social presence as ``the degree to which a person's perceptions of an agent or robot shape social interaction with that robot''. This concept is then broken down into two classes of design: ``embodiment'' and ``co-location''. Each class has attributes for creating rich, social interactions; in this paper we focus on exploring the physical embodiment of artificial agents.

The \textit{embodiment hypothesis} in socially interactive robotics \citep{wainer2006role} argues that \textbf{a robot's physical presence augments its ability to generate rich communication}. The physical embodiment of social agents provides them with more modes of communication that can be used to convey internal states and intentions in more intuitive, human-like ways \citep{lohan2010does}. \citet{barsalou2003social} concisely outlined four significant ways in which physical embodiment directly effects the social capabilities of these interactive systems:
\begin{quote}
First, perceived social stimuli do not just produce cognitive states, they \textit{produce bodily states} as well. Second, perceiving bodily states in others \textit{produces bodily mimicry in the self}. Third, bodily states in the self \textit{produce affective states}. Fourth, the compatibility of bodily states and cognitive states \textit{modulates performance effectiveness}.
\end{quote}

\subsubsection{Types of Socially Interactive Robots}

Socially interactive robots vary in many aspects of embodiment and social ability. They can be classified into seven categories according to \citet{fong2003survey}, expanding on \citet{breazeal2003toward}:
\begin{enumerate}
\item \textbf{Socially Evocative:} Robots that evoke feelings stemming from the natural human tendency to nurture and care for anthropomorphized agents.
\item \textbf{Social Interface:} Robots that use social cues and communication modalities familiar to human users. This requires embodiments to be capable of generating (and often also understanding) those social cues.
\item \textbf{Socially Receptive:} Robots that are socially passive but benefit through interaction. They are limited in the social cues they are capable of learning by their respective embodiments.
\item \textbf{Sociable:} Robots that proactively interact with humans to complete internal goals.
\item \textbf{Socially Situated:} Robots in a social environment that they are capable of understanding and reacting to \citep{dautenhahn2002design}.
\item \textbf{Socially Embedded:} Robots that are socially situated but also structurally coupled with their environment and have knowledge of human interactional structures.
\item \textbf{Socially Intelligent:} Robots that have human-level social intellect. This is be the most complex and technologically-capable class of socially interactive robots.
\end{enumerate}

\section{Summary}
In this section we discussed the bodies of work surrounding the concept of embodiment that are relevant to the field of socially interactive robots. Embodiment has been studied by a wide variety of disciplines. Philosophers have examined embodiment as a lens to the human experience, studied its relationship to human cognition, and discussed how it serves as a source for both physical and cognitive human social expectations. Psychologists and communications theorists have long been fascinated by the notion of \textit{presence} and how symbolic representation of agents can be appropriately designed. Embodiment is inherently contextual; consequently, the latest developments in communication technology and media, such as virtual reality and on-screen characters, have had a strong influence on recent studies. Since human-robot interaction is a relatively young are of robotics, the value of embodiment in social HRI is also a relatively new area of study. Traditionally, the physical embodiment of robots has been discussed in the context of functional value--perception, mobility, and manipulation. We discussed how the field has now advanced to include considerations of interactive value and design affordances. We then discussed the related fields of virtual agents, collaborative robots, and service robots, all of which have relevant aspects of embodiment.  Finally, we introduced the embodiment hypothesis that is fundamental to human-robot interaction.

{Researchers of embodiment have various opinions on its role in an artificial agents' abilities--some see limited benefits from physical embodiment \citep{hoffmann2013investigating} while others claim that ``intelligence cannot merely exist in the form of an abstract algorithm but requires a physical instantiation, a body'' \citep{pfeifer2001understanding}.  In the next section, we review a set of embodiment studies conducted in socially interactive robotics over the last decade and a half, in order to evaluate the state of the embodiment hypothesis today.}

\chapter{The Design Space for Socially Interactive Robots}\label{chap:design-space}
In design practice, formally defining a \textit{design space}---the elements that designers can vary to create possible variations in the appearance, behavior, and overall makeup of a system---can facilitate constructive discussion and systematic experimentation. Robots are complex interactive systems, and defining a design space for them can guide future development and serve as a framework for understanding prior research and identifying gaps in our knowledge. The exploration of design space for socially interactive robots typically involves \textit{industrial design}, \textit{animation}, and \textit{interaction design} to create variations in the physical construction, behavior, and interactive capabilities of robot systems. The physical design and appearance of socially interactive robots are inextricably tied to behavioral capabilities and interactivity, as they set user expectations and mental models regarding functional and social abilities of the robot \cite{goetz2003matching,lee2005human}. Furthermore, a set of contextual factors, such as the features of the \textit{task} in which users are expected to interact with the robot and the \textit{role} that the robot is envisioned to play in the task, shape user perceptions of the physical and behavioral characteristics of the robot and their interaction with it. Therefore, the design space for socially interactive robots must be defined in a way that integrates physical, behavioral, interactive, and contextual factors.

\section{Contextual Factors}

\begin{figure}
  \centering
    \includegraphics[width=\textwidth]{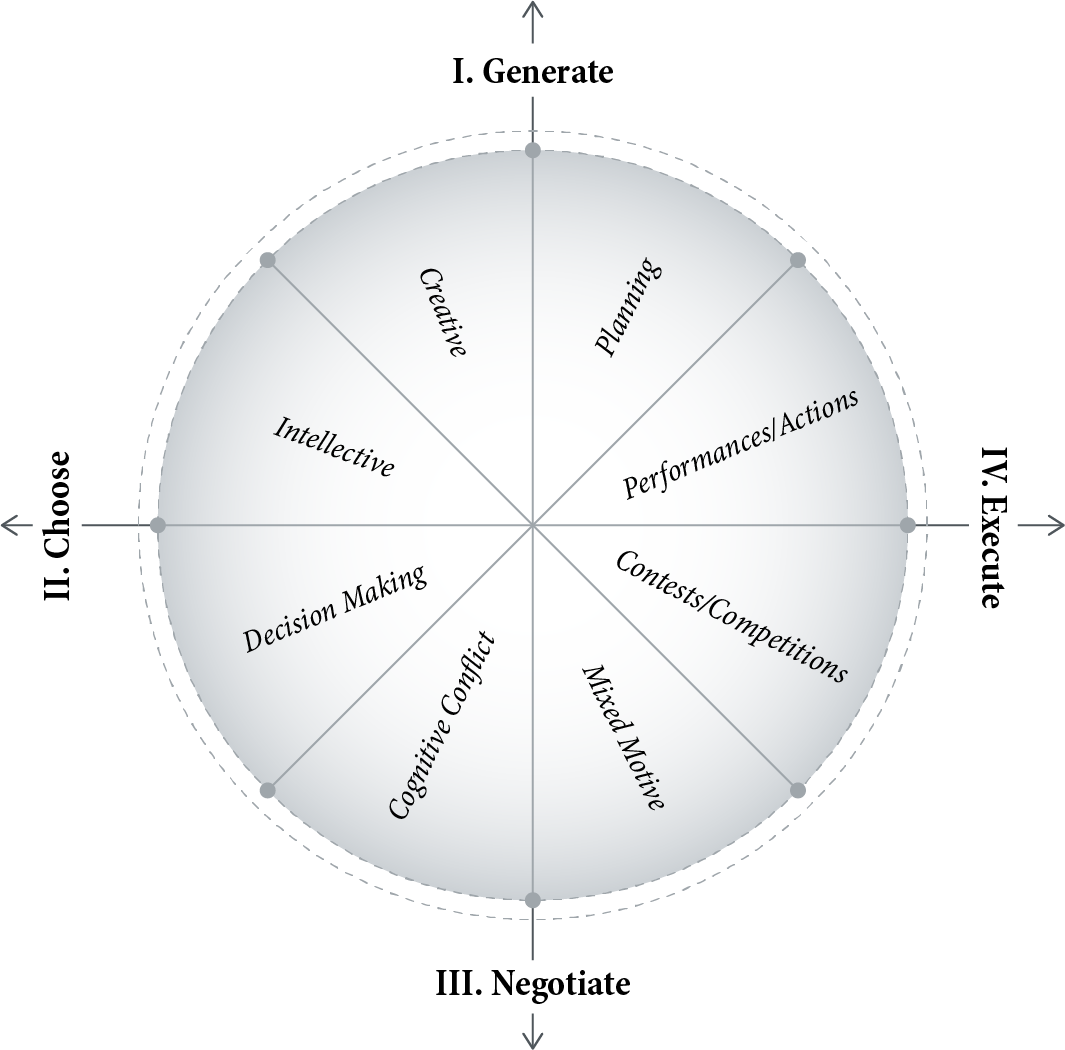}
  \caption{McGrath's \citep{mcgrath1995methodology} Circumplex of Group Tasks, segmented into octants along the dimensions of \textit{generate}-\textit{negotiate} and \textit{execute}-\textit{choose}.}
  \label{fig:taskrole-template}
\end{figure}

Interaction, whether between humans or between humans and robots, is always shaped by \textit{context}. In the following subsections, we introduce a taxonomy for the two core contextual factors of interaction with socially interactive robots: the \textit{tasks} in which robots are used and the \textit{roles} that the robots play in those tasks.

\subsection{Tasks}
\label{s-tasks} % another label
The first of the two dimensions of social context is the \textit{task} at hand. Using the Circumplex Model for group tasks proposed by \citet{mcgrath1995methodology} (seen in Figure \ref{fig:taskrole-template}), we classify the reviewed studies into {one of eight octants} along two main dimensions of (1) \textit{generate}-\textit{negotiate} and (2) \textit{execute}-\textit{choose}. The task at hand acts as the underlying driver of these interactions and is the more general of the two contextual factors being considered. Below, we define these eight task categories and provide example studies of each from our review.

\begin{itemize}
	\item \textit{Planning (Generate-Execute)}: Planning tasks are those in which a series of steps is determined by the interacting agents in order to reach a goal for the group.  For example,  \citet{vossen2009social} compared the influence of feedback relative to energy consumption used by an embodied robot and a computer. Users were asked to use a simulated washing machine interface to clean clothes while trying to minimize electricity consumption.
	
	\item \textit{Performances/Action (Execute-Generate)}: Performance or action tasks are those in which some or all members of the interaction group execute a series of actions, typically following a set of predetermined instructions, to achieve a goal. The action(s) taken depend on the task context, but the tasks share the characteristic of having \emph{quantifiable} performance metrics. For example,  \citet{bainbridge2011benefits} evaluated the use of a physically or virtually embodied artificial agent in instructing participants to perform tasks that ranged from moving stacks of books from one shelf to another to discarding stacks of books by placing them into a trash can.
	
	\item \textit{Contests/Competition (Execute-Negotiate)}: Contest tasks involve conflicts of power between interaction agents in action-based, competitive tasks \citep{posner2005circumplex}.  The competitive components involve negative-sum or zero-sum games \citep{nash1951non} and thus associate negative cost (both social and functional) with task-related decisions. For example,  \citet{bartneck2003interacting} asked participants to compete against robot agents in a negotiation task involving stamps that were assigned values prior to the start of the game. Both the robot and the participant were trying to maximize their individual scores and could negotiate and trade with one another throughout the activity.
	
	\item \textit{Mixed-motive (Negotiate-Execute)}: Mixed motive tasks involve resolving conflicts of interest among interacting agents \citep{posner2005circumplex}. Such conflicts are structured as positive sum games, in which the net benefits received by an individual party do not necessarily detract from the benefits of another \citep{nash1951non}.  For example, \citet{shinozawa2003robots} explored the use of robotic agents in retail settings with relevant social goals such as conversing with participants about purchasing a set of kitchen knives.
	
	\item \textit{Cognitive conflict (negotiate-choose)}: Cognitive conflict tasks involve resolving conflicts of viewpoints among interacting agents \citep{posner2005circumplex}. For example, \citet{pereira2008icat} used the iCat robot to play chess with participants, starting from a predetermined mid-game position with the participant at a slight advantage.
	
	\item \textit{Decision Making (Choose-Negotiate)}: Decision-making tasks are those in which interacting agents
	decide issues with no unique correct answer \citep{posner2005circumplex}. For example, \citet{lee2015impact} asked participants to rate the ``genuineness'' of smiles in artificial agents and robots by comparing Duchenne and non-Duchenne smiles.
	
	\item \textit{Intellective (Choose-Generate)}: Intellective tasks are similar to decision making tasks but have correct answers. For example, \citet{zlotowski2010comparison} had participants solve math problems with the robot agent as a medium for feedback related to the task.
	
	\item \textit{Creative (generate-choose)}: Creative tasks involve generating ideas. While painting, composing, and photography are examples of typical creative tasks, in the context of \citet{mcgrath1995methodology} task circumplex, those tasks are classified as \textit{performance} tasks given that actions are being taken. In contrast, \citet{fischer2012levels} is an example of a creative task that asked participants to describe objects to the robot that were selected by the experimenter.
\end{itemize}

\subsection{Social Roles}
\begin{figure}
  \centering
    \includegraphics[width=\textwidth]{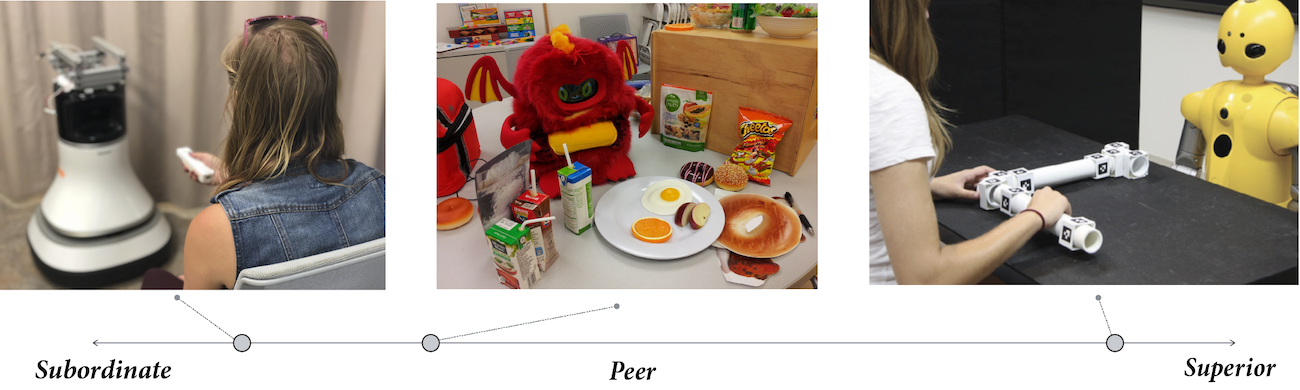}
  \caption{Examples of robot applications in different social roles, including a \textit{subordinate} mobile base following remote controls, a \textit{peer} eating ``buddy'' for children, and a \textit{superior} robot ``instructor'' that gives the user task directions.}
  \label{fig:social-roles}
\end{figure}

The second dimension of context is the {\it role} the agent plays in the interaction. Roles are inherently tied to the agent's abilities to achieve certain contextualized goals, both social and task-oriented. For example, agents in the role of a ``superior'' may be capable of delivering trustworthy information and gaining adherence because of their perceived reliability and competence \citep{kennedy2015robot}; those perceived as ``peers'' may facilitate interesting and engaging cognitive competition; and ``subordinate'' agents may improve user self-efficacy and encourage attachment formation through a balance of demonstrated ability and disclosed incompetence \citep{bartneck2003interacting}. Understanding how people respond to agents of varying social roles is critical for designing socially interactive robots. To discuss these roles, we look to classical works in organizational theory \citep{magee20088}, plotting roles of agents along the spectrum that spans from subordinate to superior (Figure \ref{fig:social-roles}).

\section{Design Paradigms}
The second defining feature of socially interactive agents is the design of their {\it embodiments}, or their \textit{industrial design}. The form that the agent's embodiment takes --- physical, virtual, or disembodied --- and the potential benefits of that form are key design considerations. Some researchers have characterized different forms of embodiment along the ``weak'' to ``strong'' axis \cite{duffy2000intelligent}. \citet{mutlu17virtual} argued that the choice of virtual or physical representation goes beyond a weak vs. strong sense of embodiment to elicit disparate frames of min d and result in vastly different user experiences. Virtual embodiments bring users into the agent's environment, invite them to participate in a crafted narrative, provide proxemic relationships that are constrained and determined by physical arrangements and conventions, and offer a safe setting to experience emotions. Physical embodiments, on the other hand, are co-situated in the users' environment, perceived as independent agents pursuing their own goals, and seen as real-world, self-relevant stimuli. Interactions with physical embodiments emerge through joint action and intention, and proxemic relationships with these agents are dynamic and co-managed to follow human norms \citep{mead2016perceptual,mead2017autonomous}.  Despite these significant differences in the \textit{nature} of interactions with virtual and physical embodiments, the body of work that we review here considers the form of embodiment to be a design choice and seeks to establish the differences in interaction outcomes through direct comparison.

\begin{figure}
  \centering
    \includegraphics[width=0.8\textwidth]{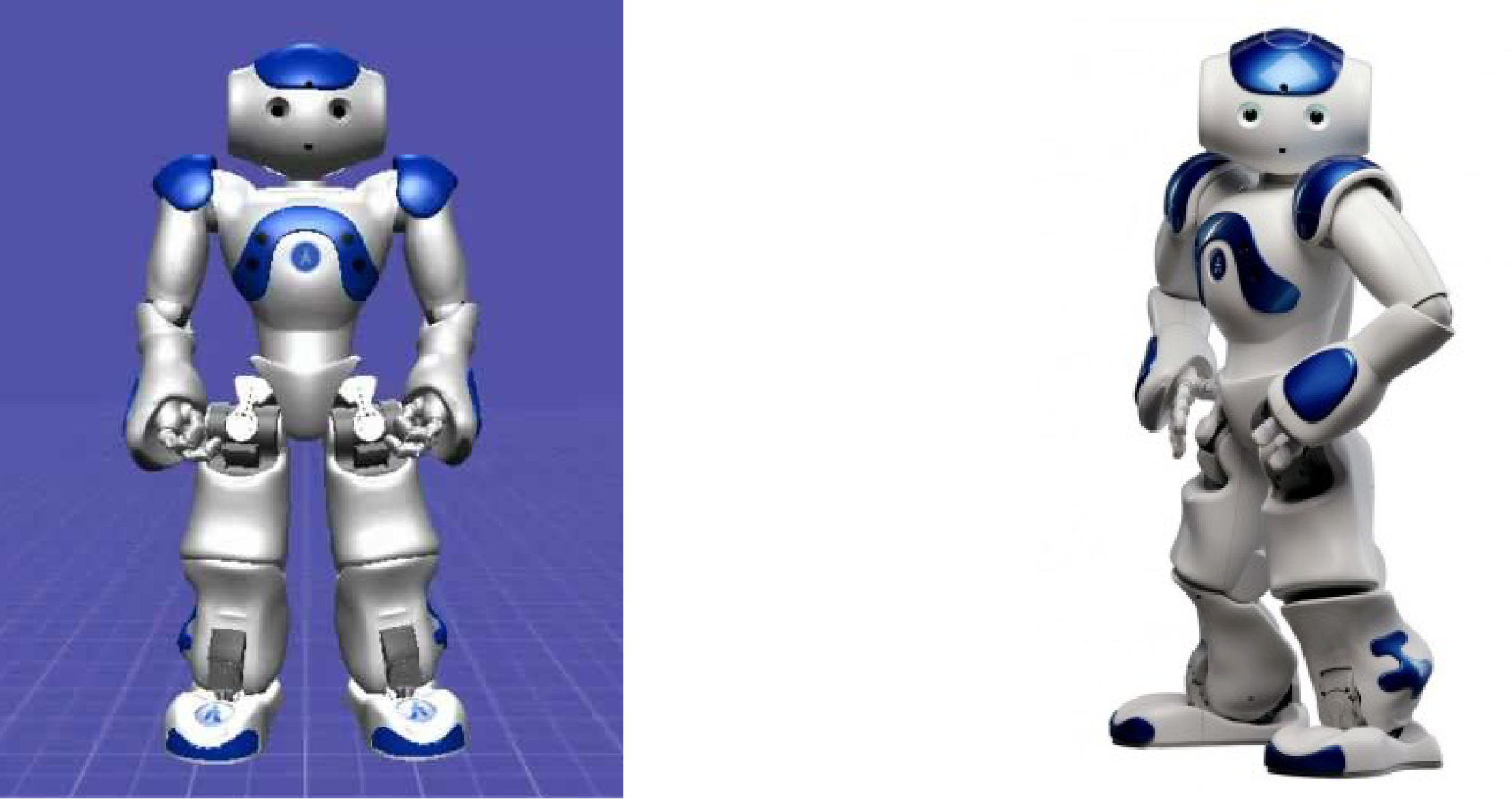}
  \caption{A virtual NAO robot (left) and a physical NAO robot (right), representing the virtual and physical or \textit{weak} and \textit{strong} embodiments.}
  \label{fig:virtual-physical}
\end{figure}

\begin{figure}
  \centering
    \includegraphics[width=\textwidth]{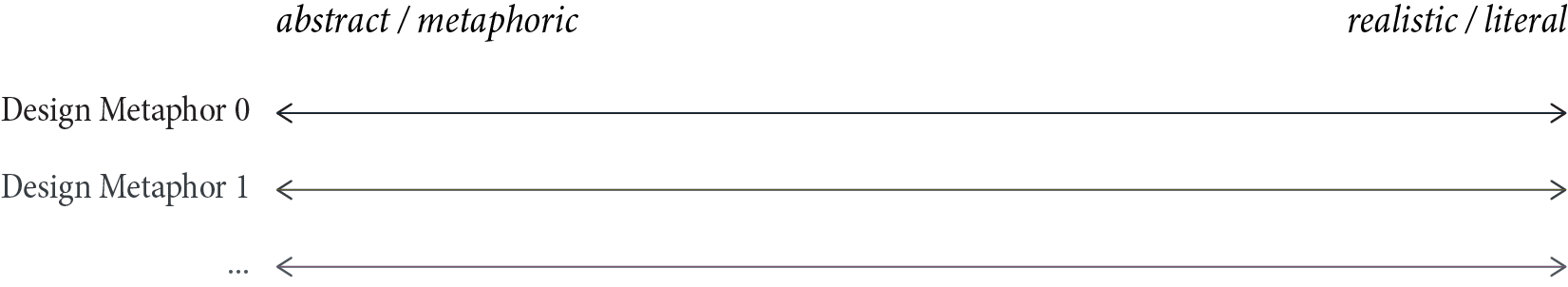}
  \caption{A characterization of the design of embodiments for artificial agents. Designs follow discrete \textit{metaphors} but vary along a continuous axis of \textit{abstraction}.}
  \label{fig:metaphor-abstraction}
\end{figure}

Comparing physically embodied robots to their virtual counterparts (Figure \ref{fig:virtual-physical}) to test the value of physical embodiment in artificial agents is a common theme in the reviewed research.  However, research on embodiment in socially interactive robotics also attempts to learn about specific design features and methods that may be used to create more engaging and effective robot systems. The design space for the embodiment of robots and virtual agents is vast. Because there are so many features of a robot's embodiment, the robots and virtual agents used in the reviewed studies vary greatly in their designs. To address the variability of embodiment design, we focus on two dimensions of every robot's design: (1) \textit{design metaphor} and (2) \textit{level of abstraction}, characterized in Figure \ref{fig:metaphor-abstraction}.

\subsection{Design Metaphors}
The notion of the \textit{design metaphor} stems from traditional design fields and refers to the design inspiration of an artifact, or in our case, robot. The metaphor for a robot's embodiment affords certain expectations for interaction partners and scaffolds social interactions. For instance, a humanoid robot with a mouth is more likely to be expected to speak compared to a bird-like robot with a beak. The design metaphors for socially interactive robots cover a wide range of possibilities, including cats, dogs, people, and cars. Since there is no simple linear relationships between these different metaphors (i.e., the metaphor of a cat is not obviously somewhere between the metaphor for a dog and a human), we define this subset of the robot design space as a discrete, nonlinear space. Because the design of embodiments can be inspired by multiple metaphors, we classify each embodiment by its primary design metaphor and discuss its level of abstraction relative to that singular design metaphor.

\subsection{Abstraction and Stylization}
\begin{figure}
  \centering
    \includegraphics[width=\textwidth]{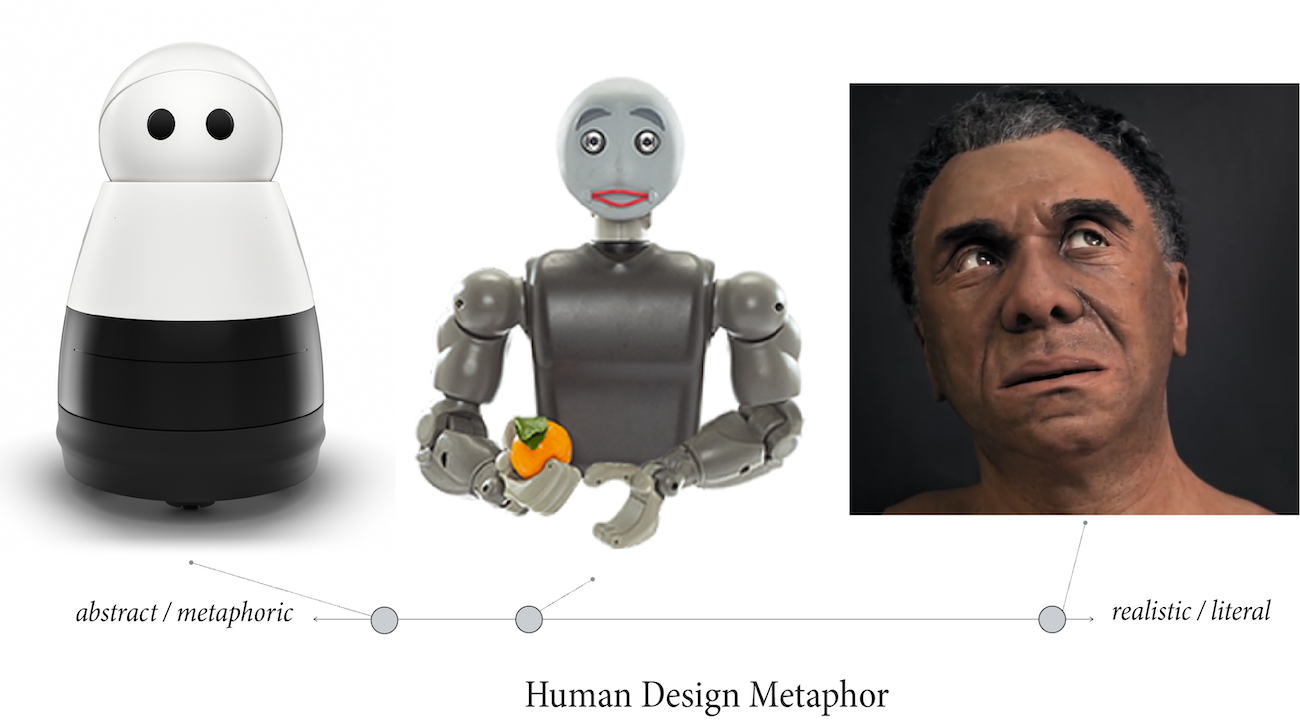}
  \caption{Three example robot embodiments spanning the spectrum of abstraction for the anthropomorphic/human design metaphor: Kuri (left), Bandit (middle), and Engineered Art's Mesmer (right).}
  \label{fig:level-of-abstraction}
\end{figure}

The {\it level of abstraction} at which the design metaphor is manifested on the robot's embodiment defines how known abilities and characteristics from the design metaphor elicit expectations about the robot's capabilities. An example of differences in abstraction for the same design metaphor can be seen in Figure \ref{fig:level-of-abstraction}; all three robots are inspired by the human form and inherit varied subsets of human embodiment features such as arms, eyes, and mouth. The robot on the left, Kuri, looks much more abstract than the robot in the middle, Bandit, and the robot on the right, Mesmer, is much more human-realistic than the other two. Because of the differences in abstraction of their human-inspired forms, perceptions of these robots will differ and can affect the performance of each robot in different task scenarios.

\section{Behavior Design}
\begin{figure}
  \centering
    \includegraphics[width=\textwidth]{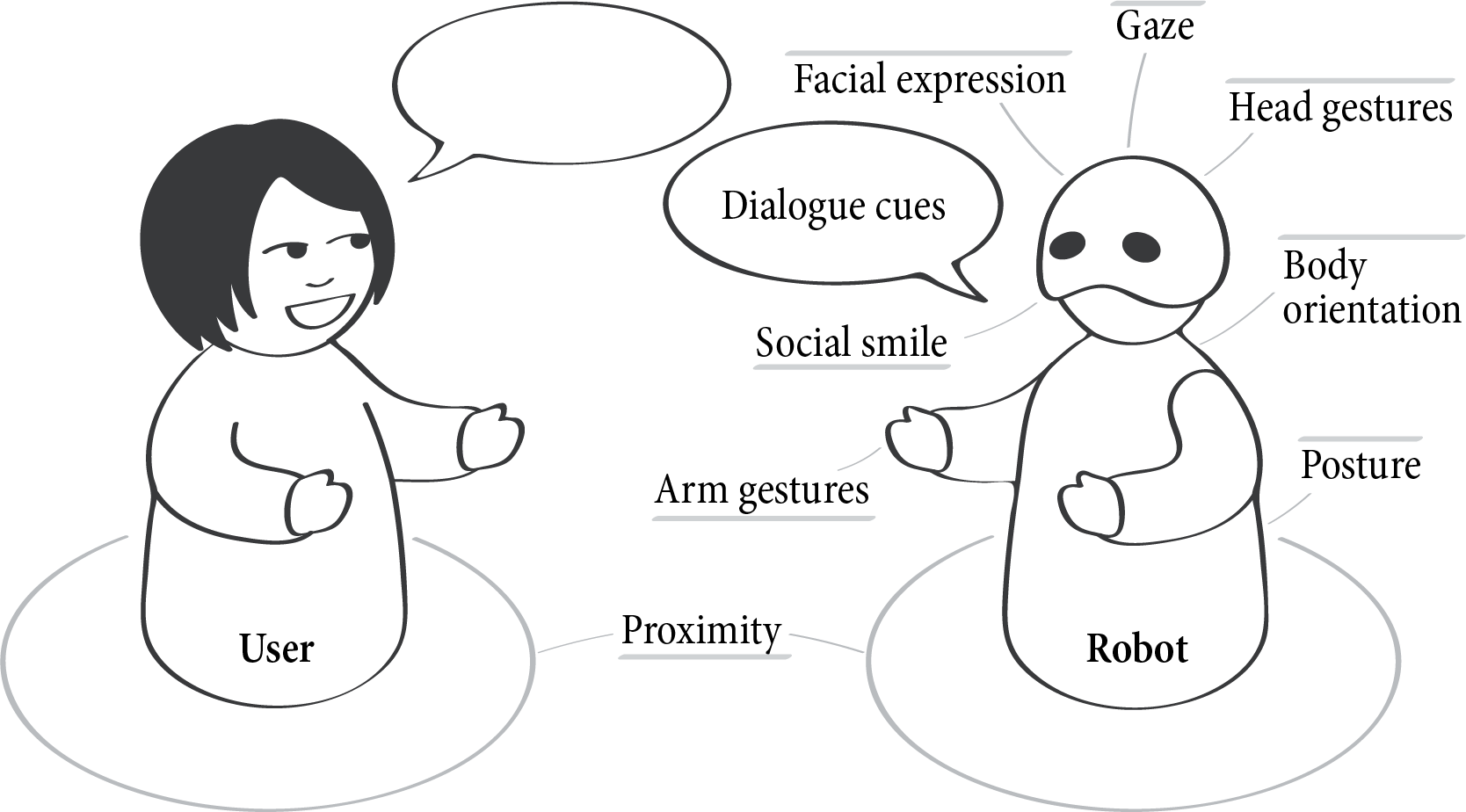}
  \caption{Some of the \textit{behavior} design variables for socially interactive robots (adapted from \citep{mutlu2011designing}).}
  \label{fig:bilge3i}
\end{figure}

Human-robot interaction is grounded in human-human interaction and multi-modal communication patterns. Embodied robots can leverage rich channels of communication that are unavailable to purely text- or speech-based interactive systems. Work in human communication has provided evidence that embodied interaction, when effectively executed, can elicit improved performance in various social, cognitive, and task outcomes \citep{antle2009lifelong,lee2012effects}. In embodied interaction, agents utilize behavioral mechanisms that encompass both the ability to perform specific behavioral elements and the timing with which these behaviors are used in the context of the interaction. We refer to these behavioral elements as {\it embodied cues} \cite{mutlu2011designing}. This section provides an overview of embodied cues used by agents in the reviewed studies or explored in human communication, focusing on cues that we believe will be important design variables for creating effective socially interactive robots.

\subsection{Limb-Based Gestures}
A large subset of embodied cues consists of hand, arm, and head movements, which---when used according to the norms of human\break communication---can communicate a wide range of ideas and create rich and salient interactions \citep{krauss1998we,mcclave2000linguistic,Deng-2018-1}. Those limb-based gestures fall into five primary categories: \textit{iconic}, \textit{metaphoric}, \textit{beat}, \textit{cohesive}, and \textit{deictic} gestures \citep{mcneill2008gesture}.

\textit{Iconic gestures} are used to communicate ideas directly related to the semantics of the associated speech, while \textit{metaphoric gestures} are used to communicate more abstract concepts and ideas.  Both use ``pictorial representations'' commonly expressed through hand and arm movements \citep{krauss1998we}. Iconic gestures range from sign languages, which explicitly and specifically convey assigned semantics of the communicator's messages, to more general gestures that convey less specific meaning, such as ``large.'' \textit{Beat gestures} are related to physical representations of prosody and pace of speech and are often used to emphasize specific segments in speech and to maintain timing and pace during the interaction. These gestures can involve a wide variety of motions, including repetitive hand, arm, head, or full-body movements. \textit{Cohesive gestures} are used to associate thematically related segments of speech and improve coherence and clarity of speech. Using similar gestures at targeted points during a verbal presentation helps observers to construct relationships between ideas being presented. \textit{Deictic gestures}, or pointing gestures, are used to provide references and direct attention toward objects in the shared environment. These gestures are performed with arm, hand, or head movements and serve as cues for establishing joint attention in situated interactions.

\begin{figure}
  \centering
    \includegraphics[width=\textwidth]{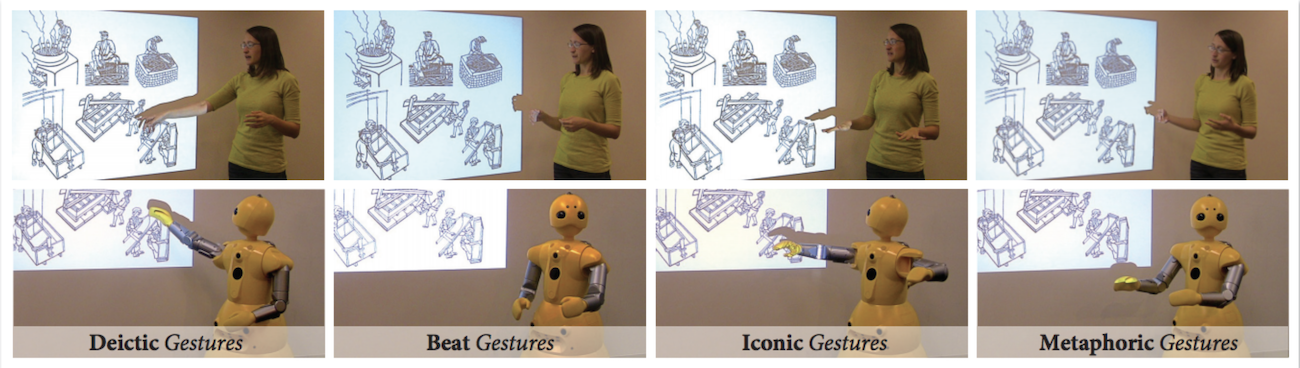}
  \caption{Examples of various limb-based gestures of embodied robots from work by \citet{huang2013modeling}.}
  \label{fig:bilge3ii}
\end{figure}

Head movements commonly serve as deictic, cohesive, and beat gestures.  Speakers use them in the form of pointing or directional cues; listeners use them in the form of nods and shakes \citep{mutlu2011designing} to signal understanding and agreement, as back-channel cues for attention and uncertainty, and as a means of pacing interactions \citep{Deng-2017-1}.

\subsection{Posture}
The embodied cues discussed above are explicitly performed using different combinations of embodied features at specific times during an interaction \citep{knapp2013nonverbal,tomasello2010origins}. The overall poses of the agent's body  in its ``resting state'' are also important, as they provide cues about attitude and status relationships in the interaction \citep{mehrabian1969significance}. By observing the overall orientation and the kinematic configuration of the agent, researchers have shown significant correlations between posture and speech, allowing prediction of upcoming speech from video \citep{mcquown1971natural}. Posture cues, such as the ``arms-akimbo position'' where a communicator places the hands on the hips and bows the elbows outwardly, convey information about the internal state of the communicator, shaping how others perceive the communicator \citep{osborn1996beauty,mutlu2011designing}. Researchers have studied these phenomena and developed systems that enable socially interactive robots to better interpret, and therefore generate, explicit posture cues \citep{gaschler2012social}. These outcomes highlight the importance of posture in the embodiment design space.

Posture, like limb-based gestures, are particularly affected by differences in robot embodiments. Different robot hardware inherently constrains embodied expressive gestures in different ways; mapping semantic gestures across different forms of embodiments is an important open challenge for generalizable expressions for socially interactive robots \citep{wang2006developmental,tosun2014general}.

\subsection{Gaze}
The gaze cues of an individual, defined by the orientation---and shifts thereof---of the eyes, the head, and the body, convey rich information about the direction of attention and mental and emotional states of the individual \cite{frischen2007gaze}. These cues serve a range of social functions, including facilitating turn-taking \cite{duncan1972some,mutlu2012conversational}, helping to establish joint attention \cite{emery2000eyes}, and signaling the intent and mental states of others \cite{calder2002reading,byom2013theory,huang2015using}. The wide range of functions that gaze serves is due largely to their highly contextualized nature. For example, the aversion of gaze during turn-taking can help speakers to more effectively manage conversational roles, while listeners can use gaze aversion to regulate intimacy and put the speaker at ease \cite{andrist2013conversational,andrist2014conversational}. Gaze cues also serve as a supplement to or a replacement for deictic gestures \cite{sato2009commonalities}, providing speakers with the ability to direct attention toward objects in the environment \cite{frischen2007gaze}, and to disambiguate what is being referred to in the environment \cite{hanna2007speakers,huang2012robot}. Through gaze cues, individuals can signal personality \cite{andrist2015look}, mental states \cite{calder2002reading}, and affect \cite{mason2005look}. When used effectively, these cues can significantly enhance interaction outcomes, such as improved recall of information \citep{mutlu2006storytelling,andrist2012designing}, management of the conversational floor \cite{andrist2013conversational,andrist2014conversational}, and efficiency in task collaboration \cite{andrist2017looking}. Finally, how the eyes, the head, and the body are configured affects the perception and outcomes of gaze cues \cite{hietanen1999does,andrist2012designing,pejsa2015gaze}, highlighting the complexity of the role of gaze in social perception and the richness of the design space for gaze as an embodied cue in human-machine interaction.

\subsection{Facial Expressions}	
Embodied agents have the opportunity to use a variety of facial features and expressions. Facial expressions can appear alongside other embodied cues or as isolated behaviors \citep{goffman1959presentation}; they strongly influence how an agent is perceived. While the complexity of faces makes them a rich and expressive channel of communication, this also makes the design of expressions challenging. Inappropriate expressions can result in strongly negative interaction performance, such as eliciting  the Uncanny Valley phenomenon \citep{mori1970uncanny} that relates high-fidelity realism to the agent's features to perceived ``creepiness''. Furthermore, facial expressions that are incongruent with speech can confuse interaction partners \citep{mower2009human}.

\begin{figure}
  \centering
    \includegraphics[width=\textwidth]{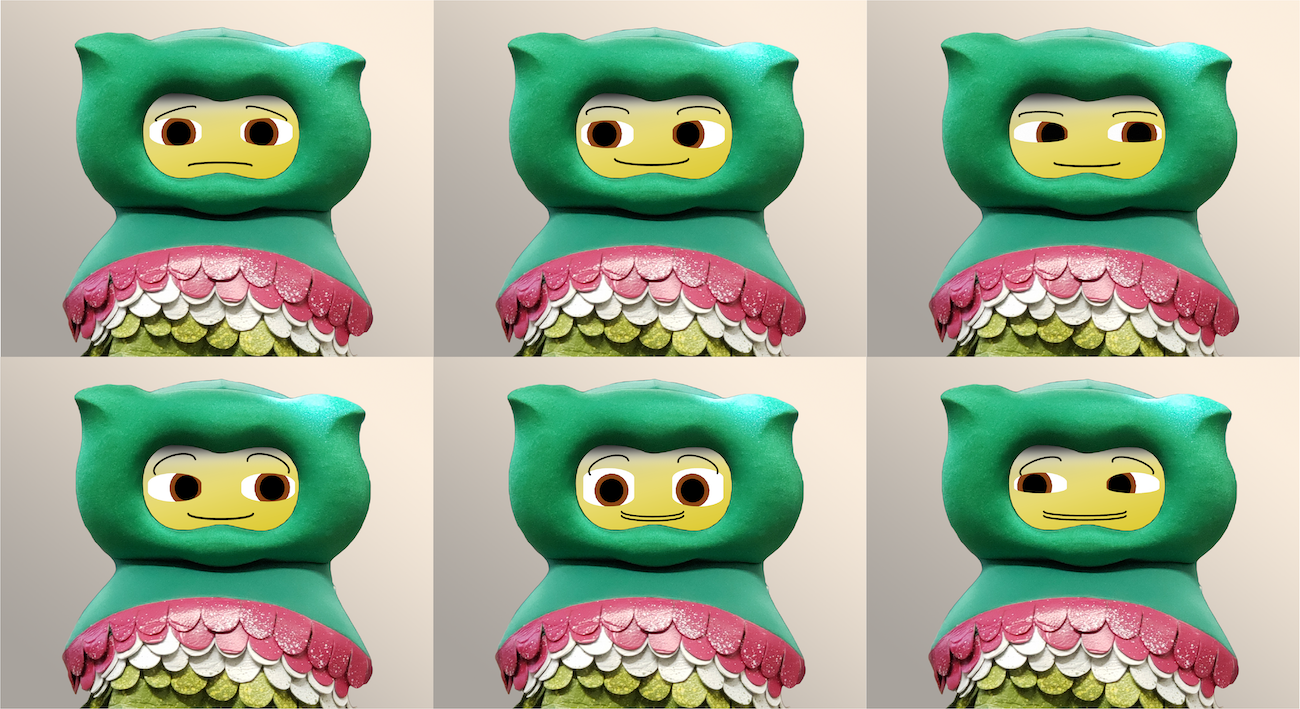}
  \caption{Different emotional expressions on the Spritebot platform inspired by expressions of human emotion.}
  \label{fig:kiwi-emotions}
\end{figure}

Social smiles are particularly important in social interaction. They serve as salient back-channel cues that express understanding and agreement, improving conversational efficiency \citep{brunner1979smiles} and perceived social competence of the robot \citep{argyle1988bodily,otta1994effect,mutlu2011designing}. Ill-timed or inappropriate smiles, however, have a strongly negative impact on interaction and can invoke the Uncanny Valley phenomenon.

Facial expressions influence the internal states of both the agent and the observer \citep{ekman1975pictures,russell2003facial}, and emotional expression and interpretation are associated with the activation of specific brain regions \citep{barrett2006solving,barrett2006emotions}. This relationship provides an opportunity for informed design for interaction.  Findings by \citet{ekman1975pictures} provided abstractions, such as ``happy'' and ``sad,'' that serve as the most commonly used foundation for the design of facial expressions for anthropomorphic and zoomorphic robots. Such robots are often designed with faces that are more abstract than the rest of their bodies relative to their design metaphor in order to enable the effective use of facial expressions and to eliminate unnecessary complexity \citep{Short-2017-998}. Zoomorphic robot designs can also utilize human ``facial action units,'' allowing designers to use human-like facial expressions on animal-like robots to express interpretable emotion \citep{Short-2017-998,kalegina2018characterizing}. This technique effectively blends animal-like and human-like metaphors as the primary and secondary metaphors, respectively, as people are much less familiar with animal facial expressions. The design of the Spritebot platform is an example of this approach, blending feline and human metaphors in the design of the robot (Figure \ref{fig:kiwi-emotions}) \citep{Short-2017-998}.

\subsection{Proxemics}
The positioning of social agents in physical space relative to other interaction partners and objects also acts as a salient embodied cue in social interaction \citep{mead2013automated}. The distance and orientation of interaction agents provide strong bidirectional signals for perception, intent, and attitude that are especially relevant for the design and implementation of mobile socially interactive robots \citep{mumm2011human,mead2017autonomous}. Research in human communication has long studied human proxemics, offering a number of models to predict how spatial behaviors affect interaction outcomes \citep{argyle1965eye,hayduk1980personal}. Work in human-robot interaction has provided experimental support for some of these models \citep{mumm2011human} and has highlighted the importance of proxemic cues in the design of interactive behaviors for physically co-present robots \citep{takayama2009influences,walters2005influence}.
The design of these cues can drastically change how people perceive robots, e.g., as disruptive and threatening \citep{mutlu2008robots} or as accepting and friendly \citep{mead2016perceptual}, underlining the need for careful consideration of proxemic behavior design.

\begin{figure}
  \centering
    \includegraphics[width=\textwidth]{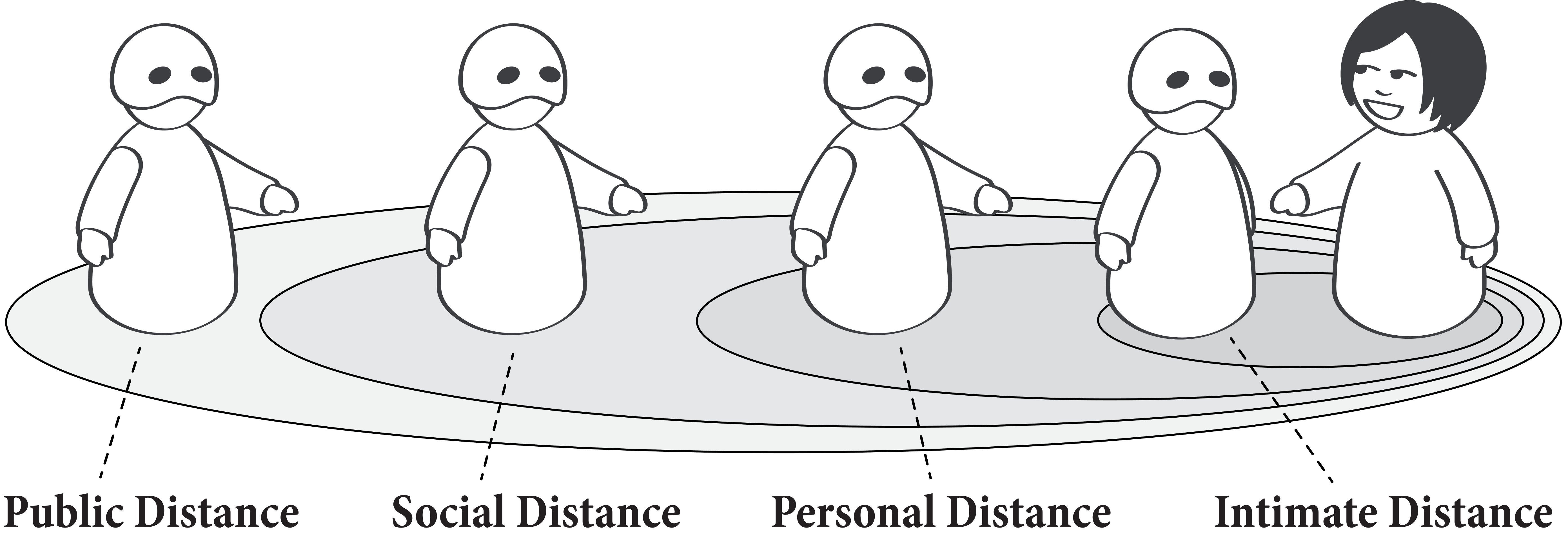}
  \caption{An illustration of proxemics zones suggested by \citet{hall1963system}.}
  \label{fig:bilge3iii}
\end{figure}

\subsection{Social Touch}
Physical embodiment presents robots with the opportunity to physically interact with their environments, including their interaction partners. \textit{Social touch} comprises non-functional touch-based interactions such as hand-holding or touches on the arm, shoulder, and face \citep{jones1985naturalistic,gallace2010science}. In human-human interaction, social touch facilitates development, social connectivity, and emotional support, and helps communicators to establish and maintain engagement throughout interaction  \citep{yohanan2012role,jung2017first,gallace2010science}. When used according to human social norms, social touch cues can serve as salient signals for dominance, intimacy, immediacy, and trust \citep{mehrabian1972nonverbal,montagu1979human,burgoon1991relational}.

\section{Summary}
This section introduced a characterization of the design space for socially interactive robots. Since design spaces have served as  transformational tools in many design fields, our goal was to provide designers and researchers such a tool for embodied interactive agents. As socially interactive robots are complex systems,  we analyzed their design aspects within three sub-systems chosen to parallel \textit{industrial design}, \textit{interaction design}, and \textit{animation}.

The embodied cues discussed in this section make up the primary elements in the design space of interactive behaviors for socially interactive robots. When designed carefully and used within established social norms, such behaviors can enable rich, engaging, and effective interactions. The next section introduces the metrics used to discuss different facets of interaction performance, reviews results from the surveyed studies, and discusses the implications of those results on the design of the behaviors, embodiments, and interaction strategies of future socially interactive robots.

% ------------------------------------------------

\chapter{Embodiment Study Outcomes and Design Implications}
\label{c-designspace} % a label for the chapter, to refer to it later
This section introduces an overview of the various evaluation techniques used in the reviewed studies, grouping them into categories of measures and discussing their strengths, weaknesses, and domains of application. This categorization of evaluation techniques and the design space taxonomy from the previous section are then used to analyze results from the reviewed studies and present their design implications.

\section{Experimental Overview}
\begin{figure}
  \centering
    \includegraphics[width=\textwidth]{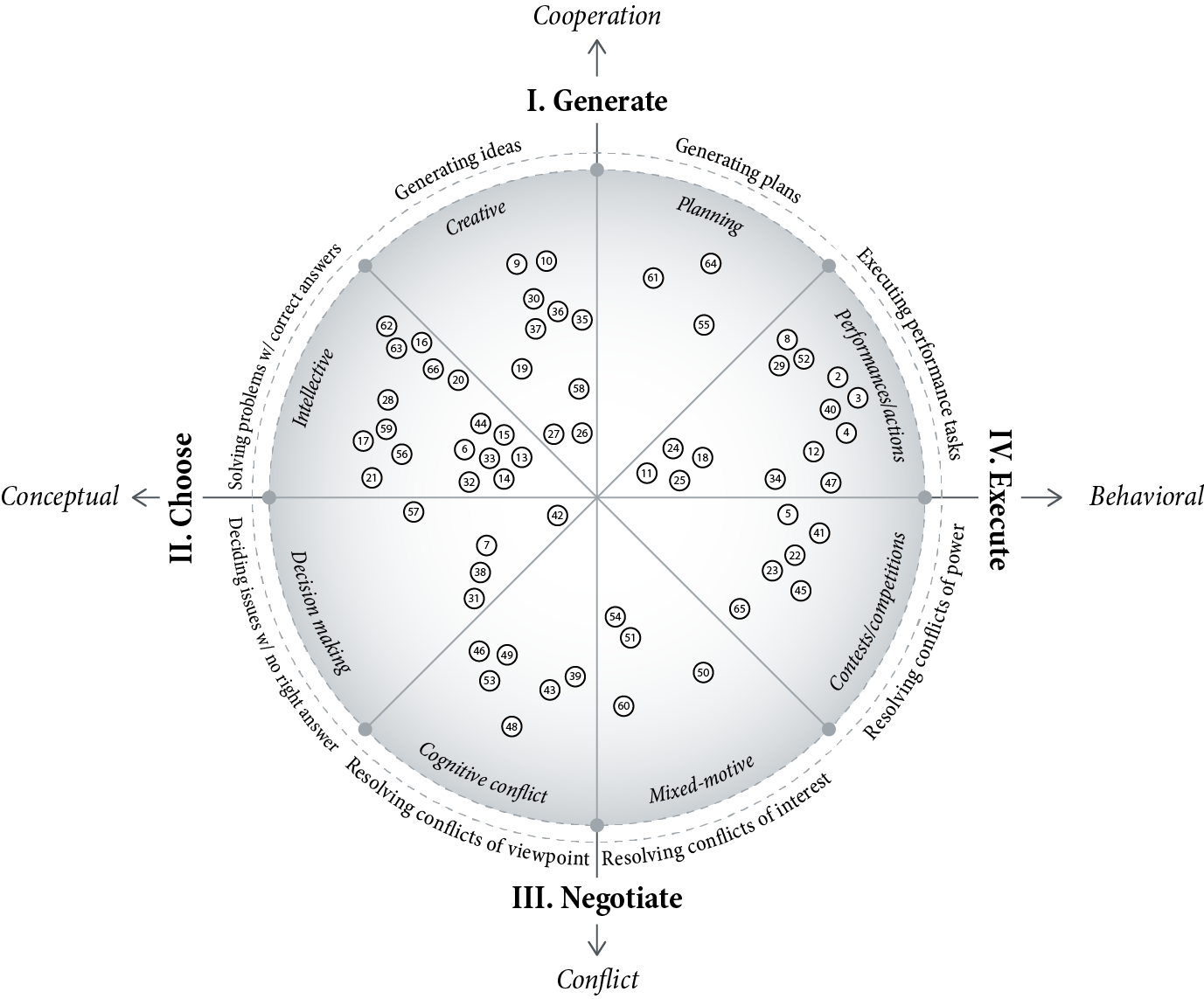}
  \caption{Studies included in our review, overlaid on McGrath's \citep{mcgrath1995methodology} task circumplex. The distance from the center indicates social role: inward = subordinate, outward = superior.}
  \label{fig:taskrole-basic}
\end{figure}

Our review of prior studies on embodiment in socially interactive robots covers a wide range of applications, user populations, and methodologies. We begin by introducing the set of experiments that we evaluated and discuss the overall landscape of embodiment studies at the time of this review. Of the 65 experiments in our review that compared a physically embodied or \textit{strongly embodied} agent to a comparable virtually embodied or \textit{weakly embodied} agent, 50 experiments compared two types of embodiments, 11 experiments compared three types of embodiments, and 4 experiments compared more than three different types of agent embodiment. Of the 65 total experiments, 17 involved more than 60 participants, 24 involved between 30 and 60 participants, and 24 involved fewer than 30 participants (Table \ref{table:population}). In Figure \ref{fig:taskrole-basic}, the reviewed experiments are mapped on the task circumplex \citep{mcgrath1995methodology}. The social role of the robot is represented by the distance from the center: the closer to the center, the more subordinate and the further from the center, the more superior.

\section{Interaction Outcomes and Measures}
\label{s-measures} % another label
As discussed in earlier sections, studies of embodiment in the humanities and social sciences predate research on the embodiment of robots and artificial agents. Some of the techniques used by researchers in those fields have been adopted into robotics-related research. Validated observational instruments, such as the POMS survey \citep{garau2005responses}, the semantic-differential scale \citep{snider1969semantic}, and selected quantitative techniques from \citet{mosteller1954selected}, have been implemented in a number of studies related to embodiment in robotics \citep{goodrich2007human}. These measurement tools are especially valuable for evaluating the more subjective results of experiments involving socially interactive robots and can provide valuable insight into the state of the embodiment hypothesis.
For instance, \citet{lee2004presence} provided empirical evidence for the mediating role of presence in people's social responses to synthesized voices; Experiment~1 described by \citet{lee2006physically} showed that people evaluated both the physically embodied agent and the interaction they had with it more positively and characterized physical embodiment as ``an effective tool to increase the social presence of an object.''

\begin{figure}
  \centering
    \includegraphics[width=\textwidth]{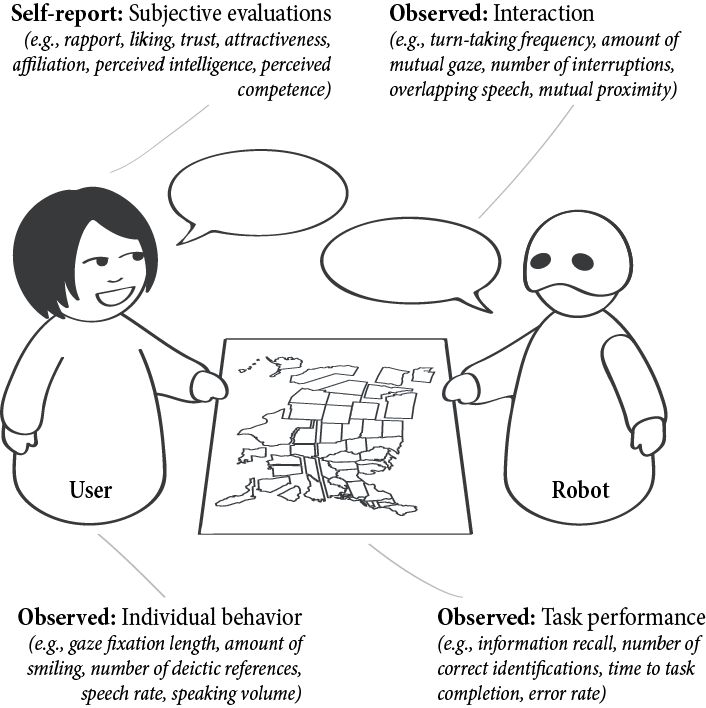}
  \caption{Measures of human experience with socially interactive robots.}
  \label{fig:measures}
\end{figure}

In previous sections, we described the relevant design elements of socially interactive agents and discussed how they can affect the \textit{quality} of user interaction with robot systems across different tasks in various social contexts. A remaining challenge is how interaction quality is defined and measured. Although prior research on embodiment captured a large number of dimensions of interaction quality, we classify measures that are used to capture these dimensions into two categories: \textit{behavioral (or observed) measures} and \textit{subjective (or self-reported) measures} \citep{goodrich2007human}. Figure \ref{fig:measures} illustrates these categories. Most studies use a combination of the two and specific measures for each reviewed experiment can be found in Table \ref{table:measures}). In the following subsections, we discuss the different measures used in the reviewed studies to provide an overview of how embodiment studies assess interaction quality.

\subsection{Self-Reported Metrics}
Self-reported measures are metrics of interaction quality collected from study participants in the form of responses to structured, semi-structured, and open-ended survey instruments. These measures give researchers the ability to capture interaction quality as perceived by participants and are especially helpful in differentiating the various facets of user experience with the robot, such as the participant's perceptions of the robot's intelligence, how much trust was established between the user and the robot, and how enjoyable participants found the interaction to be. Example self-reported measures used in prior work include open-ended interviews \citep{ju2010animate,takeuchi2006comparison} and questionnaires designed to capture various dimensions of interaction quality, including social attraction \citep{mccroskey1974measurement}, perceived intelligence \citep{krogsager2014backchannel}, and story appreciation \citep{costa2016emotional}.
Table \ref{tab:self-report-new} provides a full list of the self-report measures used in the reviewed studies.

\newcommand{\Small}{\fontsize{9.75}{11.75pt}\selectfont}

\newcommand{\Tiny}{\fontsize{5.85}{8pt}\selectfont}

%T4.1
%\begin{longtable}
\begin{table*}[p]
\centering
\caption{Measures and instruments used to capture participant perceptions of robots in the reviewed studies.\label{tab:self-report-new}}
\Small
\begin{tabular}{llc}
{\textbf Instrument} & {\textbf Measure} & {\textbf Reference}\\
\hline
User Acceptance of Information & Acceptance & \citep{venkatesh2003user} \\
Technology (UTAUT)  &&\\
\hline
Positive and Negative   & Affective State & \citep{watson1988development} \\
Affect Schedule (PANAS)&&\\
\hline
Godspeed Questionnaire & Anthropomorphism,  & \citep{bartneck2009measurement} \\
 & Animacy, Likeability, &\\
 & Perceived Intelligence, &\\
 & Perceived Safety &\\
\hline
Animated Character and  & Anxiety, & \citep{rickenberg2000effects} \\
Interface Evaluation & Task Performance,  &  \\
 & Liking &  \\
\hline
Negative Attitudes towards  & Attitude, & \citep{nomura2006altered} \\
Robots Scale (NARS) & Perceived Presence & \\
\hline
Questionnaire for & Cognitive Development & \citep{fridin2014embodied} \\
Placement Committees & & \\
\hline
NASA Task Load & Cognitive Load & \citep{hart2006nasa} \\
Index Questionnaire & & \\
\hline
Cognitive Load Questionnaire & Cognitive Load & \citep{sweller1988cognitive}\\
 & &  \\
\hline
Self Assessment Manikin & Emotional State & \citep{bradley1994measuring} \\
and Semantic Differential & & \\
\hline
Hoonhout Enjoyability & Enjoyability & \citep{hoonhout2002development} \\
Scale & & \\
\hline
Adjective-Based & Enjoyability & N/A \\
Rating & & \\
\hline
Likert-Scale Evaluations & General & \citep{likert1932technique} \\
\hline
UCLA Loneliness Scale & Loneliness & \citep{russell1996ucla} \\
\hline
Standardized Mini-Mental & Mental State, & \citep{crum1993population} \\
State Examination & Development & \\
\hline
Kidd and Breazeal Questionnaire & Perceived Presence & \citep{kidd2004effect} \\
\hline
Interactive Experiences & Perceived Presence & \citep{lombard2000measuring}\\
Questionnaire & & \\
\hline
Eysenck Personality Questionnaire & Personality & \citep{francis1992development} \\
\hline
Big Five Questionnaire & Personality & \citep{caprara1993big} \\
\hline
``I'm Sorry Dave'' Questionnaire & Sociability & \citep{takayama2009influences} \\
\hline
Children's Social Behavior & Social Behaviors, & \citep{hartman2006refinement} \\
Questionnaire (CSBQ) & Empathetic Abilities & \\
\hline
Networked Minds Questionnaire & Social Presence & \citep{biocca2003toward} \\
of Social Presence & &\\
\end{tabular}
\end{table*}
%\end{longtable}

A key consideration in the use of self-reported measures is the type of data to be collected from participants. Semi-structured and open-ended interview methods provide rich, qualitative data, while questionnaire-based survey instruments, structured using rating scales such as the Likert scale \citep{allen2007likert}, provide quantitative measurements of specific variables. For guidelines on designing interview questions and questionnaire-based measures, see \citet{louise1994collecting} and \citet{hinkin1998brief}, respectively.

\subsection{Observed Metrics}
Observed measures capture user task-related actions, physical behaviors, and physiological responses that can be observed by human experimenters or measured using sensing instruments. These measures can be evaluated in real time or {\it post hoc} and can be viewed from three perspectives: (1) \textit{individual behavior}, (2) \textit{interaction}, and (3) \textit{task performance}. The following paragraphs describe each perspective.

\textit{Individual behavior} involves observed measures of a user's behavioral, task, or physiological state over a period or at specific points in the interaction. Measures of user behavior include body motion \citep{noldus1991observer, chang2005kinematical}, body pose \citep{brooks2012simulation}, gaze behavior \citep{mutlu2009footing,mutlu2012conversational}, facial expressions \citep{costa2016emotional}, and linguistic verbosity \citep{fischer2012levels}. Table \ref{table:individual-behaviors} lists measures of \textit{individual behavior} used in the reviewed studies, the methods with which they were labeled or analyzed, and studies that included these measures.

\begin{table*}[t]
\centering
\caption{Measures of individual behavior used in the reviewed studies and methods for their capture and computation.\label{table:individual-behaviors}}
\Small
\begin{tabular}{p{1.5in}p{2in}p{2in}}
{\textbf Measure} & {\textbf Analysis Tool} & {\textbf Example Experiment(s)}\\
\hline
Attachment Level & Automated System & \citep{bremner2015speech} \\
 of Speech& & \\
 \hline
Response Time & Automated Systems, & \citep{bainbridge2011benefits} \\
 &  Human Annotation &  \\
\hline
Directed Gaze & Human Annotation & \citep{donahue2015investigating,ju2010animate,kidd2004effect,komatsu2010comparison,williams2013reducing} \\
\hline
Facial Affect & SHORE \citep{costa2016emotional}, & \citep{costa2016emotional,shahid2014child,williams2013reducing,Short-2017-998} \\
  & FACS Coding \citep{ekman1975pictures} Tools &  \\
\hline
Face Tracking & FaceAPI, & \citep{costa2016emotional} \\
  & Microsoft Kinect, &  \\
   & OpenFace \citep{amos2016openface} &  \\
\hline
Micro Behaviors & Human Annotation & \citep{donahue2015investigating} \\
\hline
Linguistic Verbosity/ & Human Annotation & \citep{fischer2012levels} \\
Breadth of Disclosure & & \\
\hline
Conversational  & Human Annotation & \citep{heerink2010assessing} \\
Expressiveness &&\\
\hline
Body Pose/ & Automated Systems,  & \citep{jost2014robot,brooks2012simulation} \\
Joint Positions & Microsoft Kinect, &  \\
  & RGBD Cameras, &  \\
    & Vicon Motion Capture, &  \\
& Human Annotation & \\
\hline
\end{tabular}
\end{table*}

\textit{Interaction} measures capture interactive phenomena that emerge through interaction among interaction partners. For example, while \textit{directed gaze} toward an object of interest in the environment serves as a measure of individual behavior, \textit{mutual gaze} emerges from two parties establishing and maintaining eye contact and serves as an interaction measure. Table \ref{table:interaction-new} lists the \textbf{interaction measures} we observed in reviewed experiments along with the analysis or labeling methods used and some sample experiments in which these measures were used.

\begin{table*}[t]
\centering
\caption{Interaction measures from observed measures used in reviewed studies.\label{table:interaction-new}}
\small
\begin{tabular}{p{1.5in}p{2in}p{2in}}
{\textbf Measure} & {\textbf Analysis Tool} & {\textbf Example Experiment(s)}\\
\hline
Directed Gaze  & Human Annotation, & \citep{lohan2010does} \\
Movement& Automated Systems &  \\
\hline
Mutual Gaze & Human Annotation, & \citep{lohan2010does,robins2006does} \\
& Automated Systems &  \\
\hline
Embodied  & Human Annotation & \citep{costa2016emotional,williams2013reducing,Short-2017-998} \\
 Nonverbal Gestures & & \\
\hline
Eye Contact & Human Annotation & \citep{fridin2014embodied} \\
\hline
Interactivity & Human Annotation & \citep{kidd2004effect} \\
\hline
Perceived Preference & Human Annotation & \citep{kose2009effects} \\
\hline
Engagement & Human Annotation & \citep{powers2007comparing} \\
\hline
Self-Disclosure & Human Annotation & \citep{powers2007comparing} \\
\hline
Attention Directing & Human Annotation & \citep{looije2012help} \\
Behaviors &  &  \\
\hline
Advise-Seeking & Human Annotation & \citep{pan2016comparison} \\
Behaviors &  &  \\
\hline
Social Touch & Human Annotation & \citep{robins2006does} \\
\hline
\end{tabular}
\end{table*}

Measurements of \textit{task performance} capture the effectiveness of the user or the group in performing the primary task of the interaction, such as effective learning in an educational interaction \citep{huang2013modeling} or speed of assembly by a human-robot manufacturing team \citep{pearce2018optimizing}. Most applications of socially interactive robots aim to support at least one quantifiable, task-oriented measure focused on the tasks in the given interaction context. In the majority of the reviewed experiments, researchers were observing interactions with defined task goals, such as performance in games \citep{fridin2014embodied}, negotiation \citep{bartneck2003interacting}, or imitation \citep{robins2006does}. These goals include explicit metrics of performance that can be used as a grounded measure of user behavior. Because task performance is inherently a contextual measure that is commonly specific to individual experiments, these metrics are highly varied.  Designing task performance measures for a given study is best informed by previous work (Appendix \ref{table:meta-overview}) in similar task categories.

As many existing studies in socially interactive robots explore new domains, applications and interaction scenarios, the research literature still lacks established and validated self-reported or observed measures. Additionally, the majority of  studies to date involve short-term interactions, and systems lack the ability to capture measurements over long periods.
As speech recognition, language understanding, affect recognition, activity understanding, and other relevant technologies improve and systems become increasingly robust, automated methods for behavior measurement over long periods will become the norm.

\section{Effects of Embodiment on Interaction Outcomes}
\label{s-interactionoutcomes}

\begin{figure}
  \centering
    \includegraphics[width=\textwidth]{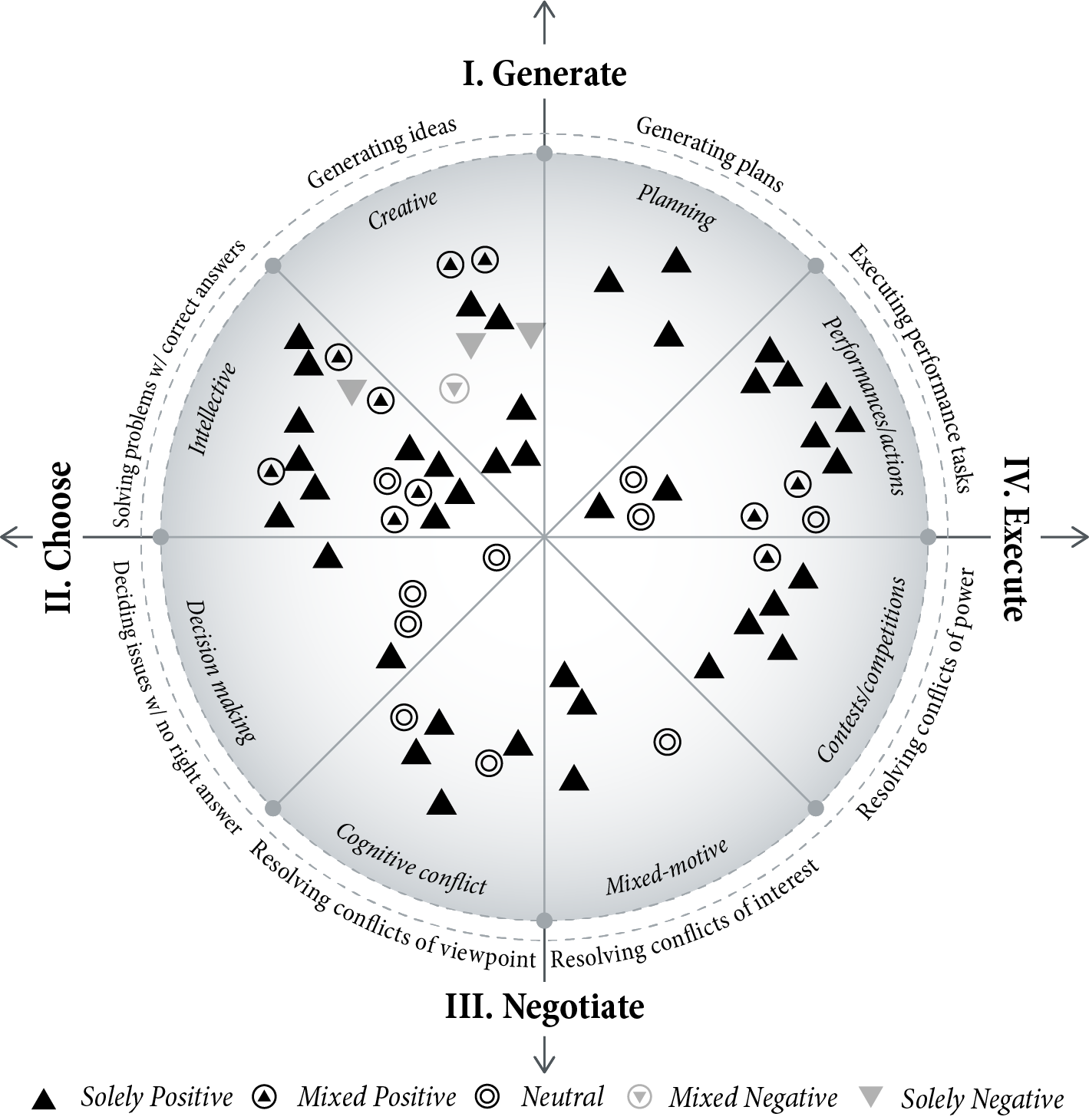}
  \caption{Combined results for all reviewed studies.}
  \label{fig:combined}
\end{figure}

The reviewed studies all seek to understand the effects of embodiment on human interaction with socially interactive robots in order to develop design guidelines for future computer and robot systems. In this section, we summarize the results of the reviewed studies with respect to this central research question.

Based on the observed and self-reported measures taken, the reviewed experiment results can be grouped into two types: \textit{differences in the perception of the agent} and \textit{differences in task performance} (Table \ref{table:results}). By combining these two measures, all experiment results can be classified into five categories relative to the embodiment hypothesis: (1) solely positive (63.1\%), (2) mixed positive (15.4\%), (3) neutral (15.4\%), (4) mixed negative (1.5\%), and (5) solely negative (4.6\%). Over all reviewed experiments, the results are strongly positive in support of physical embodiment, with 63.1\% of combined results showing improvements in interaction and performance and 6.1\% showing negative results (Figure \ref{fig:combined}).

The two measures, task performance and agent perception, are not fully separable, so analyzing both categories of measures provide a fuller and more nuanced understanding of interaction outcomes.
For instance, \citet{segura2012you} studied participants' preferences when given the option to interact with a physically-embodied robot companion or with a virtual representation of that robot. They reported that, although participants found the robot ``less annoying'' and explicitly chose to interact with it more than the virtual agent, their ratings of the different embodiments did not reflect these observed functional preferences. The authors of that work deduced that choosing between an embodied or simulated agent was very task-specific. For tasks that involve a significant amount of information transmission but relatively little social rapport (e.g., information kiosks), and for tasks that require users to reveal personal information, disembodied agents should suffice. However, for tasks that are relationship-oriented (e.g., a home companion), social engagement is important for maintaining rapport, and physical embodiment is beneficial for increasing social presence, and in turn, engagement and rapport.

In the following subsections, the results of the reviewed studies are examined across the agent performance and perception categories, and analyzed relative to the current state of the embodiment hypothesis in socially interactive robots.

\subsection{Differences in the Perception of Agent}
\label{s-percep}
The primary method for evaluating the social performance of artificial agents measures users' perceptions of those agents, and changes in those perceptions as a result of interactions. The affordances gained by the design of a robot, the behaviors of that robot, and demonstrated competence are all key components of the resulting user perceptions. Researchers have aimed to study specific features such as \textit{attachment, comfort, loneliness}, and \textit{general attitudes towards robots} \citep{bartneck2009measurement}, in isolation from other factors such as novelty effects and prior task experience.

\begin{figure}
  \centering
    \includegraphics[width=\textwidth]{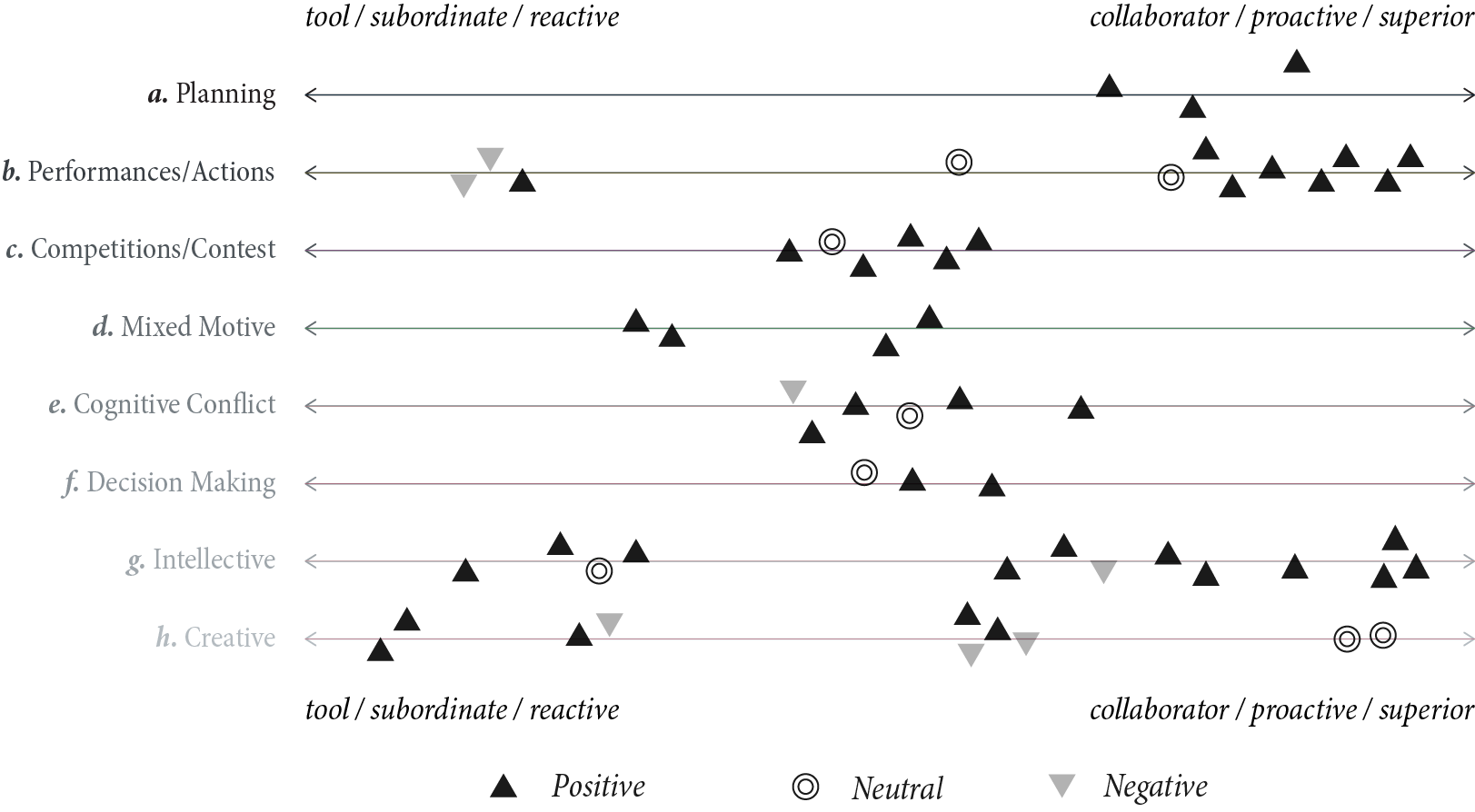}
  \caption{Survey results for \textit{interaction performance} differences between physically embodied and otherwise-embodied agents.}
  \label{fig:interaction-results}
\end{figure}

In our review, 57 of the 65 studies measured differences in the perception of the artificial agent using a variety of observational instruments. The majority of the experiments used a combination of observed and self-reported measures. The results reporting on agent perception are found in Figure \ref{fig:interaction-results} on the task circumplex. Each point represents a singular experiment, and its color represents the finding; green represents physical agents outperforming their non-physical counterparts; yellow represents a neutral result; and red represents virtual or non-embodied agents outperforming the physical agents.

Of the 57 experiments, 43 (75.4\%) showed that using a physically embodied agent is superior in improving user perceptions of the agent.  Every task category had a majority of results favoring physical embodiment; four of eight task categories had positive or neutral results. Our review provides support for the embodiment hypothesis in the context of agent perception based on the current state findings in the field.

While 43 experiments had results supporting the embodiment hypothesis, 7 experiments presented negative results, and another 7 experiments presented neutral results. Most of the neutral results (shown in yellow in Figure \ref{fig:interaction-results}) have task and role classifications comparable to experiments with supportive results (shown in green), suggesting that these neutral results may stem from other facets of experimental design and may not indicate the impact of embodiment. Because of the subjective nature of measuring agent perceptions, we postulate that the lack of statistical significance in these experiments could be due to high data variance.

The two experiments with neutral results not completely surrounded by positive results are in the \textit{creative} task category and are both the most ``superior'' agents used in their respective task categories. This pattern, along with the three negative results for the most ``peer-like'' agents in the same task category, is a strong indicator for the importance of social roles in certain types of tasks. We speculate that creative tasks may be a task category particularly sensitive to social roles, as the creative process involves socially complex interactions.

The seven negative results form clusters within their respective task categories. As Figure \ref{fig:interaction-results} shows the results of studies as a function of task type and social role, the clustering along the dimension of social roles within each type of task provides further evidence for the importance of appropriately designing social roles in the perception of artificial agents for different types of tasks.

Perception of the robot agent is affected by three factors: (1) the design of the robot (design metaphor and abstraction level), (2) the behaviors of the robot, and (3) the perceived social role of the robot. Depending on the type of interaction or task and the duration of the interaction, the relative importance of these factors can vary. The design of the robot scaffolds interactions by setting expectations about the robot, including signaling its physical and cognitive capabilities and its ability to follow norms of human social behavior. For example, a humanoid robot with a dynamic mouth is more likely to be expected to have conversational capabilities than a cat-like robot with a molded mouth. A robot with realistic-appearing arms is expected to be able to perform both gesture and manipulation tasks while a robot with stylized arms may only be expected to perform simple or high-level gestures.

People are primed by the perceived social role of the robot.  For example, users may hear out a subordinate robot but not comply with it \citep{waldron1991achieving}. Failure by a superior agent in generative tasks, or tasks that involve collaboration between the robot and person to co-create ideas or narratives, can be interpreted as incompetence, while failure in negotiation tasks can be interpreted as potentially manipulative \citep{caldwell1982responses}. When suggestions from subordinate agents fail, people may be more likely to take the blame for the failure of the suggestion---it was not the agent's incompetence that caused the failure, because the person, as the superior agent, should have known not to take that advice.

The perceived social roles and expected behaviors of robots are not static; through demonstration of their functional and social abilities, robots can show to their human interaction partners what they are capable of. The length of the interaction is particularly important factor that shapes the effects of robot design and behavior on agent perception. For example, in short-term interactions, the affordances gained by the ``first impressions'' of the robot, typically stemming from the design of its embodiment, play particularly important roles in user perceptions of that robot. In longer interactions, users are given more time to observe the behavior and demonstrated capabilities of the robot and can adjust their first impressions accordingly. For instance, if a robot is perceived to have manipulation capabilities based on its embodiment, and it fails at manipulating objects that are too heavy or too large, the perception of the robot' physical capabilities will be impacted negatively, and the expectations of the robot will become more realistic. As people's expectations are calibrated by the robot's demonstrated behavior, the affordances and impressions first gained from the embodiment become less important \citep{segura2012you}.

The robot's task is especially important in the context of \textit{perceived} social roles. In the reviewed experiments, there were some discrepancies between intended and perceived social roles and, in some cases, the social roles were themselves the experimental variables \citep{Short-2017-998}. The tasks provided contexts in which social roles could be evaluated.

%CONCLUSION
Overall, the experiments we reviewed provide strong support for the value of physical embodiment in perceived social competence, measured by people's perceptions of the artificial agents. Furthermore, our interpretations of the few negative results highlight the need for proper embodiment design. By measuring human perceptions of artificial agents, researchers aim to study their social capabilities that serve as fundamental context for \textit{task performance}, as discussed in the next section.

\subsection{Differences in Task Performance}
\label{s-perform} % another label
Agent-agent interactions typically aim to accomplish a set of shared goals. Those goals can be abstract, such as ``have a discussion'' \citep{powers2007comparing}, or explicit, such as ``move all books from current locations to goal locations'' \citep{bainbridge2011benefits}. Within the shared goals, each individual has their own goals that may represent conflict (bottom half of the task circumplex) or cooperation (top half of the task circumplex).  Compared to measurement tools used for agent perception and social performance, metrics of task performance involve observation-based, objective measures, such as measuring response time \citep{bainbridge2011benefits}, the number of moves in a puzzle \citep{hoffmann2013investigating}, or the compliance rate \citep{komatsu2010comparison}. Pairing task performance and agent-perception measures can provide a more complete understanding of interaction outcomes. Because socially interactive robots are usually designed to accomplish or support a task, even if it is as general as engaging a user for a set amount of time, task performance measures can be seen as the ``end result'' of the robot's performance.

\begin{figure}
  \centering
    \includegraphics[width=\textwidth]{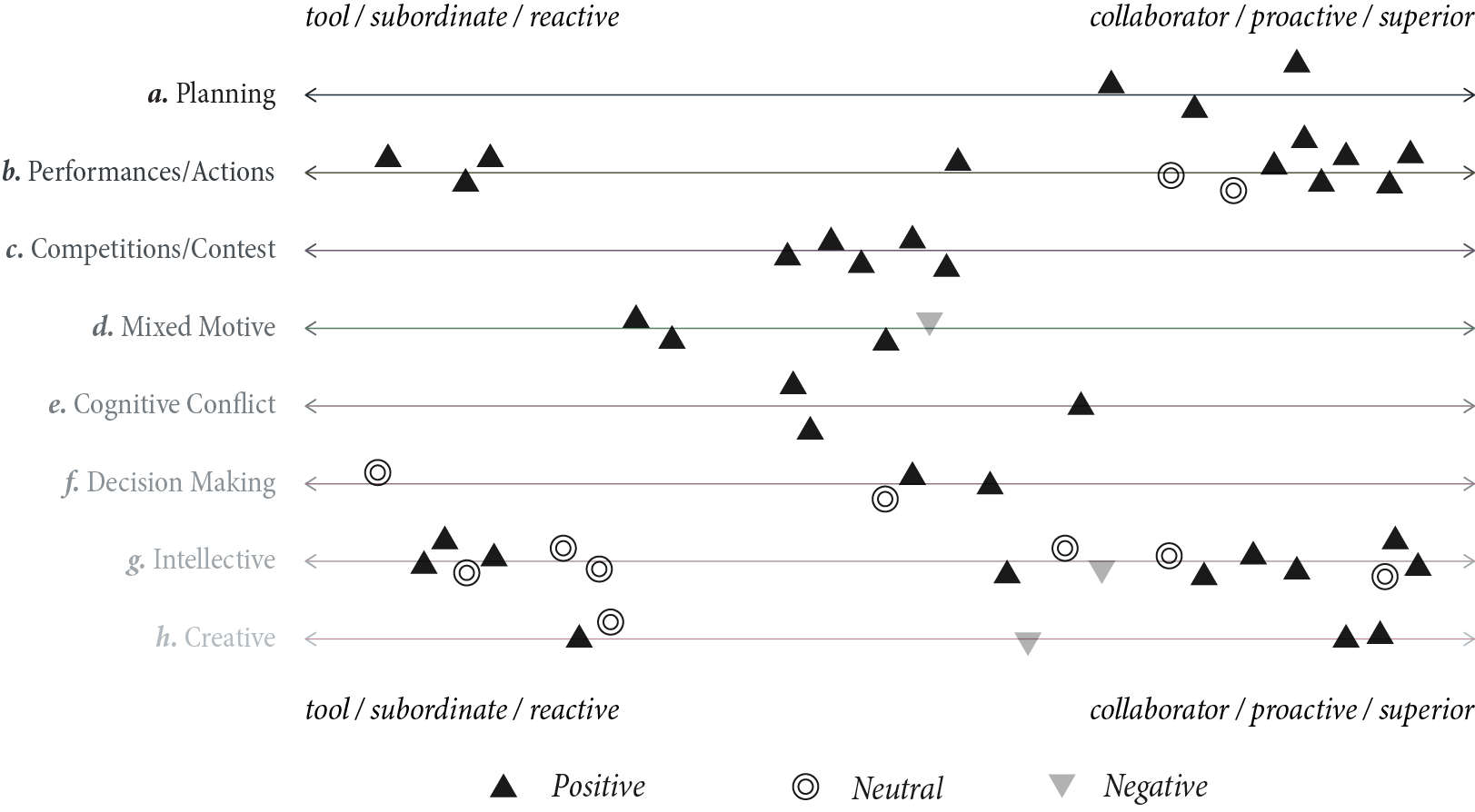}
  \caption{Survey results for \textit{task performance} differences between physically embodied and otherwise-embodied agents.}
  \label{fig:task-performance}
\end{figure}

Of the 65 studies we reviewed, 57 used defined metrics for task performance; a majority showed significant improvements in user performance when collaborating with physically embodied robots.  These task performance results, plotted over the task categories and social roles, can be seen in Figure \ref{fig:task-performance}. Of the 57 experiments, 37 presented positive results (71.15\%) for having a physically embodied robot over virtual or disembodied agent, leading to the conclusion that physical embodiment is beneficial for improving the social performance of artificial agents as well as task performance of human users interacting with those agents. For instance, \citet{jung2004effects} found that, when interacting with the eMuu robot, participants scored higher on the negotiation task than when they were interacting with a virtual rendering of the robot.  \citet{bainbridge2011benefits} presented results for a book-moving task in which the artificial agents requested unusual behaviors such as throwing books into the trash; the results showed higher compliance rate when requests came from a physical robot than when they came from a virtual agent. In learning tasks, \citet{jost2012ethological} also showed children to be significantly more motivated by a physical robot when playing a cognitive-simulation game.

Although the majority of the reviewed experiments demonstrated task performance improvements, 3 did not, and 12 had neutral results. In comparing non-positive results in agent perception and task performance, we note that their overall total numbers are similar  (14 and 15, respectively), but there are almost twice as many neutral results in task-performance measures than in agent-perception measures. The neutral results fall into four of the eight task categories and are clustered by social role within those categories. For instance, in \textit{performance and actions}, experiments show improved task performance when the embodied robot is playing either a superior or subordinate role, while all studies with neutral artificial agents have neutral results. \citet{fasola2013socially} used a humanoid robot, Bandit, to lead ``chair aerobics'' exercises with older adults in the US, and \citet{nomura2009investigation} used a robot to give directions to adults in Japan for sorting objects into boxes. The robots played similar roles and, due to the nature of these two tasks (i.e., ``generation'' and ``execution''), the peer-like role of the robots may not have inspired as much confidence in the robots' intellectual contributions as superior robots would, nor did it cause users to feel the need to support the robot as subordinate agents would.

The three negative results are all in different task categories and, within those categories, are located near the largest cluster of neutral results. This clustering may be an indicator of social roles that are less effective than others within a given type of task, making designing a robot for that application in that role more difficult.

Overall, the results of the reviewed experiments provide strong support for physical embodiment in task performance with socially interactive robots. Compared to results for agent perception, the results for task performance had fewer negative results and more neutral results. We speculate that the effects of embodiment on user task performance, which can be independent of the robot, are not large enough to reveal a statistically discernible difference, while its effects on agent perceptions, which involve evaluations directed toward the robot, may be stronger. This interpretation presents future challenges for embodiment design, e.g., designing more salient embodiments, and highlights the nuanced and complex effects of embodiment on human interaction with socially interactive robots.
The next section draws on the insights from our findings and presents design recommendations and considerations for future studies.

\chapter{Recommendations for Future Embodiment Studies}
\label{c-methconsiderations}
A wide range of behavioral research paradigms are available for studying the role of embodiment in human interaction with socially interactive robots; \citet{mcgrath1995methodology} provides a comprehensive discussion. This chapter outlines common research paradigms, study designs, independent variables, and measurements for studies of embodiment in socially interactive robotics, concluding with a discussion of limitations and open questions for such studies. We do not attempt a comprehensive survey and refer the reader to the literature on research methods for studying human interaction with technology (e.g., \citet{lazar2017research} and \citet{olson2014ways}) and on behavioral research methods in general (e.g., \citet{cozby2017methods} and \citet{price2017research}).

\section{Research Paradigms}
Research studies of embodiment in socially interactive robots to date have primarily involved controlled laboratory studies.  As socially interactive robots become more pervasive, future studies will need to consider methodological fit \citep{edmondson2007methodological} and draw on a richer set of choices under both laboratory and {\it in situ} studies.

Laboratory studies allow for a higher level of control over the variables in the phenomena being studied, at the cost of \textit{ecological validity}, the extent to which findings in the laboratory can be generalized to real-world situations. {\it In situ} studies identify representative settings, i.e., the ``field,'' of the target environment for the design of the system, introduce the system to these settings, and enable the study of human interaction with the system using comparative or naturalistic research paradigms. For example, an {\it in situ} study of an educational robot designed to improve student attention to instruction may be conducted in a real-world classroom and seek to confirm that the attentional benefits of the design can be obtained in the complex and dynamic setting of a classroom.

{\it In situ} studies are carried out in the natural setting in which a system is deployed or the target setting for which a system is designed. Naturalistic studies involve no experimental control and follow ethnographic and other field methods to capture the natural and emergent ways in which humans interact with robots. For example, \citet{mutlu2008robots} conducted a study of how workers at a hospital interacted with a delivery robot, utilizing ethnographic observations and interviews to capture behavior as well as subjective perceptions of the robot.

Comparative {\it in situ} studies, or ``field experiments,'' in contrast, involve introducing robot into a target setting, manipulating aspects of the robot's design, and using qualitative and quantitative methods to understand how these manipulations affect human interaction with the system. For example, a field study conducted by \citet{hayashi2007humanoid} introduced two socially interactive robots into a train station, varied how active and social the robots acted, and studied commuters' interactions with and perceptions of the robots. Such studies involve some control of the system's behavior or the environment (e.g., where or how the system is introduced to the setting) while allowing all other variables to vary naturally.

\section{Study Designs}
The design of a study is determined by how much control is desired, whether independent variables can be manipulated, and how many variables are considered. Studies in social human-robot interactions involve four key study designs: \textit{true experiments}, \textit{quasi experiments}, \textit{system-level evaluations}, and \textit{naturalistic studies}.

Both quasi and true experiments seek to establish causal relationships between design variables and outcomes of interactions with robots, although they differ in whether or not experimental conditions are randomly assigned to the population of interest and thus in the conclusiveness of the causal relationships identified by the study. Additionally, true experiments are most commonly used in laboratory studies, while quasi-experimental designs are most common in {\it in situ} studies.

In true experiments, participants sampled from the population are randomly assigned to study conditions that correspond to different levels of an independent variable. For example, a study on the effects of physical proximity between the robot and its user may establish ``close'' and ``far'' distances at which the robot will interact with its user, randomly assign members of its study population to these levels, and use inferential statistics to determine whether the amount of distance  had a significant effect on participant behaviors or perceptions of the robot.

Quasi experiments, on the other hand, are used in situations where random assignment is not possible, and studies compare matched groups or make pre-/post- comparisons. For example, a study that compares the use of a robot across two senior living facilities or a study that compares social interaction among members of a facility before and after the introduction of a robot. While quasi-experimental designs can allow the exploration of settings or interactions that are otherwise impossible to study and can offer valuable insights, their findings are less conclusive than those obtained in true experiments.

Both true experimental and quasi-experimental study designs have inherent limitations when used in the context of research into socially interactive robotics. First, they offer insight into relationships between a small number of design variables in isolation and lack the ability to conveniently study large design spaces that robotic systems involve. Second, they usually show that manipulations of variables significantly affect interaction outcomes but do not provide an understanding of the extent of these effects, limiting the ability to make fine-grained design decisions to meet the specific demands of a robot product. System-level study designs seek to address these limitations by simultaneously modeling the predictive relationships between a large number of design variables and interaction outcomes. Research in socially interactive robotics has utilized two variations of this approach. The first variation involves asking users to interact with a socially interactive robot in the way it is intended and ad hoc modeling of predictive relationships between design variables and interaction outcomes. For example, \citet{peltason2012talking} asked participants to perform an object-learning task with a robot and modeled the predictive relationships between design variables such as how many utterances the robot spoke per minute, as they are naturally utilized in the interaction, and interaction outcomes such as perceived ease of use of the robot using multivariate regression techniques. The second approach utilizes the same statistical modeling tools, but instead of modeling variable-outcome relationships, it explicitly manipulated multiple design variables simultaneously within their possible ranges. \citet{huang2014learning} demonstrated the use of this method in a study on the design space of arm gestures for a socially interactive robot; they manipulated the frequency of each type of arm gesture and modeled how well the use of each arm gesture predicted interaction outcomes. They found, for example, that the use of pointing gestures by the robot significantly predicted information recall in participants. Furthermore, they found that each standard deviation increase in the use of this type of gesture increased information recall by 0.123 and 0.623 standard deviations for females and males, respectively. This example illustrates the power of system-level study designs in gaining a fine-grained understanding of variable-outcome relationships, which can enable fine-grained decisions in the design of the robot system.

Finally, research into embodiment in socially interactive robotics also benefits from naturalistic studies that involve minimal levels of control. These studies utilize methods from ethnography, the systematic study of people, settings, organizations, and cultures based on observational data, and other forms of qualitative empirical research, including fly-on-the-wall observations, participant observation, interviews, and system-log data. Data obtained using these methods are analyzed rigorously using qualitative methods. Such studies seek to utilize the rich data obtained from the setting coupled with rigorous analysis to arrive at a deeper understanding of the use of the robot system in the study setting.

\section{Independent Variables}
A common characteristic of all studies of embodiment in socially interactive robotics is inquiry into the effects of system-level properties of embodiment on the interaction, including the high- and low-level variables that make up the design of the robot. In lab and {\it in situ} studies that involve experimental control determine these properties and variables {\it a priori}. Naturalistic studies, on the other hand, explore those effects in an unstructured fashion, although they can also seek to describe studied phenomena without drawing any conclusions about them. In experimental design, those properties are called {\it independent variables} and refer to factors in the study that are explicitly manipulated, such as the height of a robot, or measured, such as the age of a participant.  The interaction outcomes, such as how approachable participants find the robot, are referred to as {\it dependent variables}.  The goal of the study is to understand how the independent variables affect the dependent ones.

Embodiment studies consider system-level independent variables that include whether the system is physically embodied, virtually embodied, or disembodied (such as a speech-based interface). Other independent variables include high-level properties of the design of the robot system, such as the metaphor followed in the design of the system. For instance, \citet{hinds2004whose} compared robots designed to follow human and machine metaphors to understand the effect of human-likeness of the robot on the attributions that human collaborators made to the robot. Finally, independent variables also include low-level design variables, such as the distance a robot maintains between itself and its user, the amount of eye-contact the robot establishes with its user, and the overall height of the robot system. These low-level properties can be variables that vary on a continuous scale, such as height, distance, or frequency, or among a discrete set of options, such as the color of the robot's lips, as manipulated by \citet{powers2006advisor} and found to affect participant perceptions of the robot.

\section{Measurements}
Understanding variable-outcome relationships in studies of embodiment requires the appropriate definition of dependent variables that are expected to be affected by independent variables of interest and the appropriate measurement of those variables. These measurements, as we previously discussed, can be categorized into \textit{observed} and \textit{self-reported}.

Observed variables include physiological reactions, behaviors, interactions, and task actions that can be reliably recognized, described, and quantified by third-party human coders, sensors, and recognition algorithms. Specifically, observed task actions of participants can be translated into standardized task-performance measures, also called ``objective'' measurements, that can be used in quantitative analyses. For example, in a task in which participants collaborate with a socially interactive robot to sort toy blocks, the number of blocks sorted by the participant or the team within a period of time can be calculated from observed task actions and can serve as a measurement of task performance. Similarly, observed participant behaviors, such as gaze shifts, gestures, and speech, can be coded into behavioral variables that can serve as indicators of high-level cognitive processes. For example, the targets and timings of the gaze shifts of participants can be translated into measurements of gaze fixation toward particular types of targets, which can signal the amount of attention paid toward these targets. Measurements from observed behaviors can be automatically extracted using technology, including sensors and recognition algorithms, such as eye-tracking technology that can automatically translate gaze behaviors into gaze-fixation measurements. The use of overt physiological reactions such as body temperature as measurements, however, require the use of technology. Finally, the interdependent behaviors of multiple agents, humans and/or robots, as they unfold over time can be translated into measurements that indicate the fluency of the interaction. Examples of interaction measurements extracted from observed behaviors include rate of turn taking in conversation, the amount of mutual gaze between a robot and its user, and the mutual physical distance maintained between parties in an interaction.

While physiological responses, behaviors, and task actions can be observed and then reliably translated into quantitative measures, participant attitudes, perceptions, and experiences are only accessible through the use of self-report measurements. Common methods to obtain self-report measures include the use of validated survey instruments such as multi-item questionnaires and translating transcriptions of interview data into quantitative metrics. The development of validated survey instruments that are appropriate for socially interactive robotics research is still in its infancy, and thus research to date has adapted validated scales from other fields and domains, including social psychology, interpersonal communication, and human factors, to study user attitudes toward and perceptions and experiences with socially interaction robots or gauge the cognitive, affective, and attentional states of participants. For example, the NASA Task Load Index (TLX) \citep{hart2006nasa} is commonly used to measure user task load when interacting or collaborating with a robot. Researchers have also coded interview transcripts to extract subjective measures such as the frequency of the use of words with positive or negative valance.

Measurements in naturalistic studies primarily use qualitative data; other study designs can supplement quantitative measurements with qualitative data. Qualitative data most commonly take the form of rich narrative descriptions of studied settings and transcriptions of interviews conducted with study participants. While technology such as audio- and video-recording can be used to conveniently capture observations and interviews, commonly used methods for qualitative data analysis, such as content analysis \citep{krippendorff2004reliability} and Grounded Theory \citep{strauss1997grounded}, require textual data.

The choice of measurement can be guided by specific design goals or hypotheses. For example, a study investigating the extent to which an instructional robot improves student learning may use observed task-performance measures that indicate learning effects, such as recall of instructional material or ability to correctly apply it to a given problem. However, the scenarios and settings are usually complex, requiring researchers to use triangulation---the use of two or more measures, methods, or approaches to assess a single relationship \citep{rothbauer2008versatile}---in order to improve the validity, confidence, and conclusiveness of findings. For example, the study on the learning benefits of the instructional robot may find significant learning effects, although this benefit may come at the expense of positive student experience. Triangulation by simultaneously measuring cognitive and affective learning would enable the researcher to gain a more comprehensive understanding of interaction outcomes and reveal potential tradeoffs.

\section{Hypotheses}
Rigorous application of many of the research paradigms described above require the development and testing of a set of hypotheses on the relationships between design variables and interaction outcomes. Naturalistic studies are generally incompatible with hypothesis-testing. Controlled laboratory or \textit{in situ} studies, on the other hand, involve \textit{a priori} consideration of independent and dependent variables and thus provide necessary elements to construct testable hypotheses.

A key consideration in hypothesis development is the basis of the prediction made by the researcher. In the context of socially interactive robotics research, hypotheses are constructed by drawing from three key sources: prior research, pilot data, and design goals. Prior research may suggest understudied but plausible variable-outcome relationships or offer preliminary findings that require more conclusive evidence. Such preliminary evidence can also be obtained from pilot studies. In socially interactive robotics, design goals can also inform the development of hypotheses, as the justifications for the design choices for a system can provide sufficient bases for an expected outcome.

\section{Limitations and Open Issues}
Current paradigms and practices in studies of human interaction with socially interactive robots have a number of limitations for consideration by future theoretical and methodological advancements. A fundamental limitation is the integrated nature of robotic systems that only allows isolating specific design variables in the context of the 	 holistic design of the system. This introduces two problems. First, and most fundamentally, robotics is challenging because of a large number of factors including uncertainties and limitations of perception and action and difficulties of repeatable behavior, as well as the challenges of robust behavior and finally the costs associated with sufficient hardware for large studies.  Next, findings from studies that isolate low-level features of a robot system may not be generalizable, because these manipulations may not be representative of the abstract design element. For example, studies of gaze behavior often interchange eye gaze, head orientation, and their combined behavior based on the level of fidelity in which gaze mechanisms are designed in a robot system or the stylized representation of gaze chosen for the design. However, whether or not findings obtained from a study that manipulates head orientation to understand the effects of gaze on user attention would generalize to other forms of gaze is unknown.  Second, system-level studies of embodiment often involve comparisons across ontologically different systems that may afford  different design variables, and comparisons at the design-variable level may not be feasible. For example, a study on the effects of touch cannot compare a physical and virtual embodiment, as the latter does not afford physical touch, and techniques to simulate touch may not effectively represent natural human experience.

Another limitation that is common in studies reviewed here is the relatively small sample sizes employed and the underpowered findings that may be obtained. Several reasons underlie this limitation. First, the nascent state of robotics in general and socially interactive robotics in particular limits the ability to utilize some of the well established practices of empirical research, such as power analysis due to a lack of prior research that would aid in estimating expected effect sizes. Second, conducting studies with complex, potentially unreliable, and often prototype systems imposes a high cost on conducting large numbers of trials. Third, the complex, interdependent, and often fluid (due to technological advancements) design spaces of these systems require a large number of system-level and variable-level studies and motivate the use of small, rapid, and iterative experimentation. Finally, the domain-driven nature of the development of robot systems often requires sampling from special populations, such as individuals with social deficits due to developmental disorders or trauma, that may show high variability in behavior or characteristics, may have clinical demands such as the presence of a therapist, or may be difficult to recruit. While single-subject studies and qualitative studies are appropriate to study robot systems with these populations, these research paradigms are not widely adopted by the research community.

Studies that seek an understanding of human attitudes toward, perceptions of, and experience with robot systems still lack appropriate and reliable survey instruments for measurement. While some scales have been developed by the community, the validity and reliability of these instruments have not been established. Adaptations of instructions from other fields, including social psychology, interpersonal communication, and human factors, do not always result in appropriate or valid measures. For example, a two-item scale of ``mutual liking'' developed for studying dyads that asks the members of the dyad how much they liked their partner and think that their partner likes them and that reliably provides highly correlated results may not be appropriate in the context of studying human-robot interaction due to the ontologically asymmetrical nature of the interaction.

Finally, the use of qualitative research paradigms, methods, and measures is still rare in research in socially interactive robotics despite their potential for deeper understanding of human interaction and experience with robot systems, particularly in naturalistic settings and when quantitative methods are inappropriate. The human-computer interaction (HCI) research community has adopted and uses a wide range of qualitative paradigms and methods with success and can serve as a model for research in socially interactive robotics. Studies that are naturalistic in their entirety or those that utilize qualitative data for triangulation are essential for exploring the effects of robot systems that are integrated in human environments.

\chapter{Implications for Designing Embodiment}
As mentioned in previous sections, socially interactive robotics is a highly interdisciplinary field that draws from a broad variety of fields, including more mature fields such as product design, human-computer interaction (HCI), and mechatronics. Designers in these fields use various techniques to implement functional designs, including heuristic evaluation, user testing, critical-path analysis, and iterative design. In socially interactive robotics, designers not only have to consider the expected challenges, such as cognition, processing, perception, manipulation, locomotion, and HRI, but also the new challenges introduced by social interaction \citep{breazeal2004designing,dautenhahn1997could, fong2003survey}. \citet{fong2003survey} introduced four design issues unique to socially interactive robots.
\begin{itemize}
\item \textbf{Human-Oriented Perception:} Socially interactive robots must have the ability to actively perceive and accurately interpret human activity and behavior \citep{fong2003survey}.
\item \textbf{Natural HRI:} Socially interactive robots must display believable behaviors, establish appropriate expectations, manage social interactions with their users, and follow human social norms \citep{fong2003survey}.
\item \textbf{Readable Social Cues:} Socially interactive robots must be able to (1) provide feedback about their internal states and (2) allow humans to interact with them in a facile, transparent manner, for example, using facial expressions, body and pointing gestures, and vocalization.
\item \textbf{Real-Time Performance:} Socially interactive robots must operate at human interaction rates. Such robots need to simultaneously exhibit competent behavior, convey attention and intention, and handle social interaction.
\end{itemize}

The rapidly growing body of work in socially interactive robotics is providing data and insights to guide the selection and development of robot platforms for new studies and deployments. Regardless of whether researchers decide to use an off-the-shelf platform, modify an existing robot, or build an entirely new robot platform, the aggregated results from past studies can inform and facilitate the robot-selection and design processes.
	
\looseness=+1
The results of the reviewed studies not only highlight the benefits of physical embodiment in many different task contexts, but the aggregated evaluations also revealed patterns that can guide the design of socially interactive robots (Table \ref{table:platforms}). By first formalizing both the application and design space of these robots and then grouping experimental design and results, we aim to provide a structured process for designing socially interactive robots informed by results from prior studies.

\begin{figure}
  \centering
    \includegraphics[width=\textwidth]{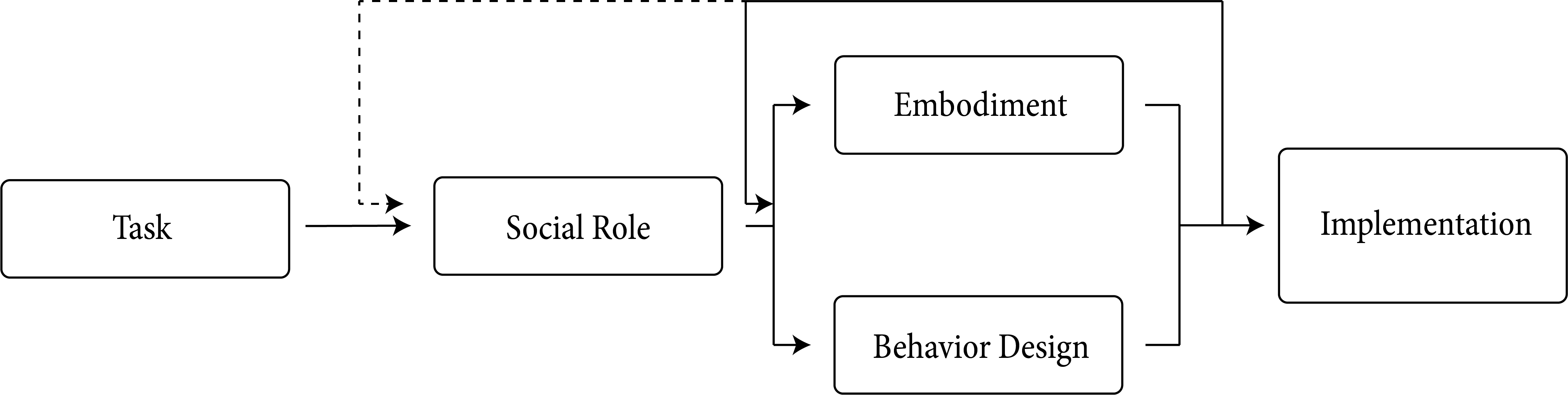}
  \caption{A characterization of the process for designing or selecting socially interactive robots for different tasks.}
  \label{fig:pipeline}
\end{figure}

\looseness=+1
Figure \ref{fig:pipeline} shows the general overview of our recommended approach for designing an application-specific robot: given a task or application, make an educated decision on what role the robot should play in that task, then select the design inspiration and set of behaviors for that robot, and then finally implement the design metaphor in the most appropriate level of abstraction. We will now discuss how we can use the results of our survey as well as existing literature in tangential fields to advise  design decisions throughout this whole pipeline.

\section{Selecting Social Roles}
\looseness=-1
We have discussed socially interactive robots and their applications in the context of (1) \textit{tasks}, (2) \textit{social roles}, and (3) \textit{embodiment}. Following Figure \ref{fig:pipeline}, we first consider the robot's task and, based on that task, select a social role for the robot. Properly selecting the role is critical because it is closely tied to a robot's ability and approach to achieving its goals; for instance, a superior robot may be a more effective teacher or coach based on heightened perceived authority \citep{bainbridge2011benefits}; a peer-like robot may be more engaging for a competitive task \citep{jost2012robot}; and a subordinate robot may help in improving self-efficacy \citep{fischer2012levels} or encouraging empathy \citep{Short-2017-998}. While selecting the and roles is still a process that requires intuition, data from the ever-growing body of past work can be used to inform the process.

\looseness=-1
Based on the reviewed studies, we can advise decisions on the social role that the robot may most effectively play in the context of that task. As the social role of a robot is a design parameter assigned by the designer, the distribution of social roles used across different task categories are a representation of the general intuition of researchers, resulting in an uneven distribution of experiments across task categories as seen in Figure \ref{fig:pipeline}. The performance of the robots playing these roles can then be used to predict potential performance of other robots for similar tasks.

\begin{figure}
  \centering
    \includegraphics[width=\textwidth]{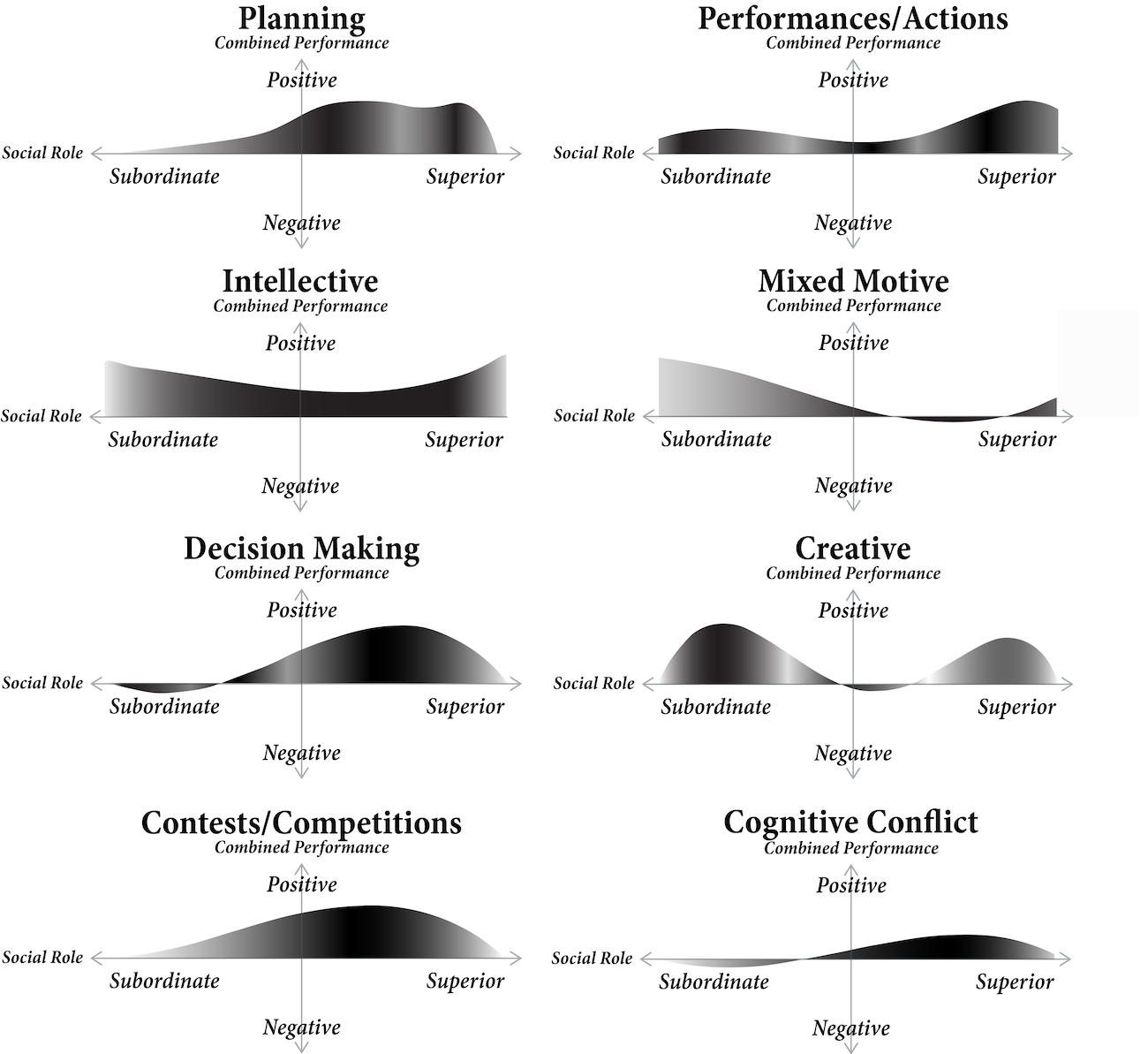}
  \caption{Visualization of artificial agents used for different task categories in the reviewed studies. Performance is \textit{combined performance} as defined in Section 4.3.}
  \label{fig:social-role}
\end{figure}

\section{Designing Robot Embodiment}
After selecting a social role to be implemented for a socially interactive robot, the designer must develop two components: the robot's embodiment and its behaviors. Both are critical for successfully implementing a robot's desired social role; in the context of this paper, we focus on the embodiment.

\begin{figure}
  \centering
    \includegraphics[width=\textwidth]{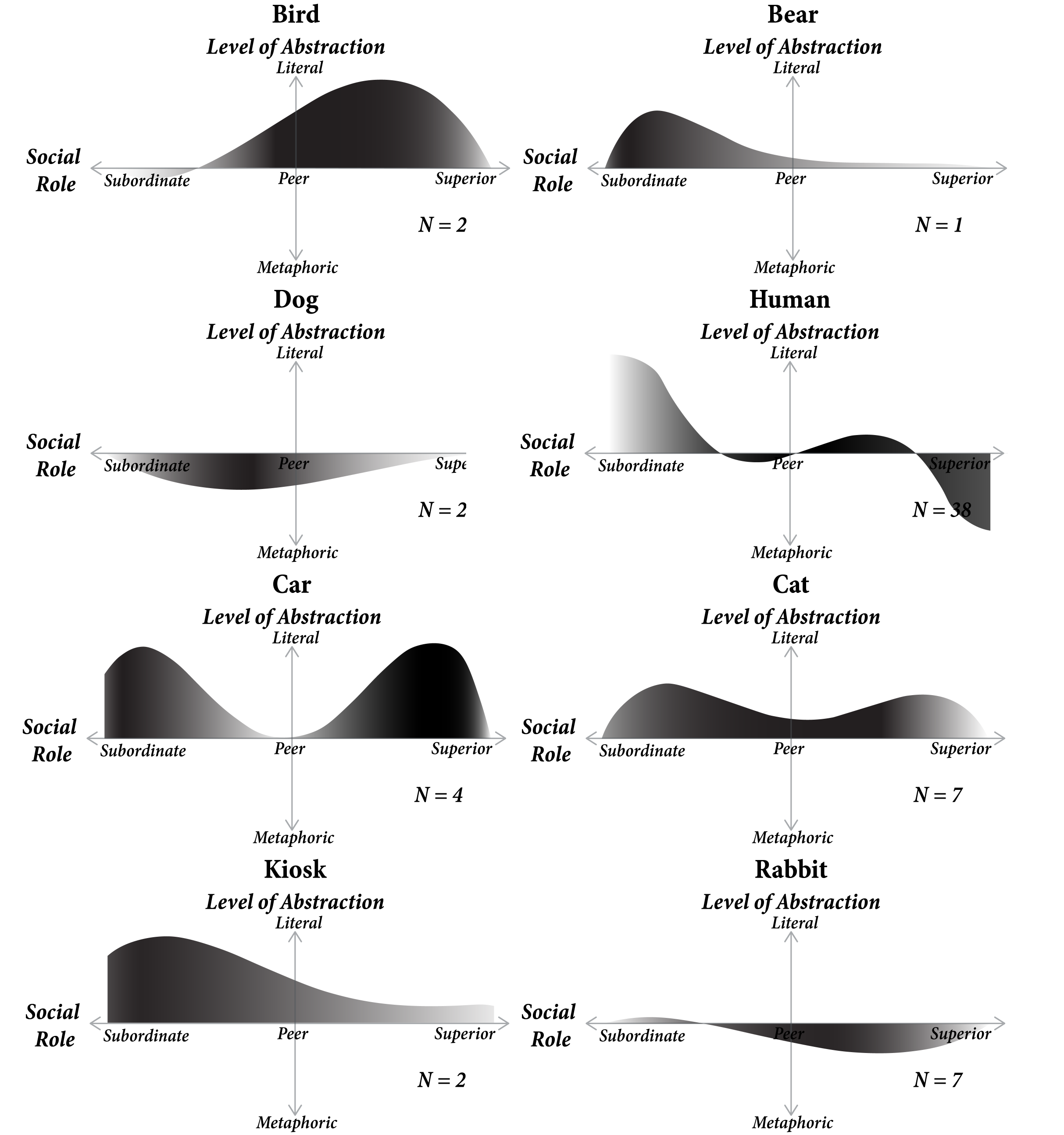}
  \caption{Visualization of the level of abstraction and social role by design metaphor for the systems in the reviewed studies.}
  \label{fig:embodiment-selection}
\end{figure}

Using our previously-discussed representation of robot embodiment consisting of \textit{design metaphor} and \textit{level of abstraction}, we can use existing studies to advise the design or selection of robots to be used in future work \citep{deng2018}. In our analysis of the 65 studies, we classified robot systems with the design metaphor and a level of abstraction (a numerical value of 1 to 10, smaller  mapping to more abstract), seen in Table \ref{table:meta-overview}. Figure \ref{fig:embodiment-selection} shows the implemented social roles of robots in the reviewed studies, for each design metaphor and by levels of abstraction.

Because of the differing mental models of the various design\break metaphors, the same social role can be implemented with different design metaphors by changing the levels of abstraction for each design metaphor. For example, if a superior social role is desired, we can reference Figure \ref{fig:embodiment-selection} to find that implementing a \textit{metaphoric human form}, a \textit{slightly literal bird form}, or a \textit{literal car form} may all effectively achieve the specific goal. Because of this flexibility, if a robot designer is constrained by either the design metaphor or the level of abstraction, they can reference existing data to advise the selection of the unconstrained design dimension and more effectively explore and iterate through the space.

\begin{table*}[t]
\centering
\caption{Robots used in reviewed studies labeled with the \textit{design metaphor} and \textit{level of abstraction }that they were assigned.\label{table:meta-robots}}
\Small
\begin{tabular}{p{1in}p{0.75in}p{0.75in}p{2.75in}}
 & {\textbf Design} & {\textbf Level of } &  \\
{\textbf Robot} & {\textbf Metaphor} & {\textbf Abstraction} & {\textbf Studies Used} \\ \hline
Keio U Robotphone            & Bear    & 4  &  \citep{ligthart2015selecting} \\
Pioneer 2DX                  & Car     & 8  &  \citep{donahue2015investigating,wainer2006role,wainer2007embodiment} \\
Pioneer P3AT                 & Car     & 8  &  \citep{segura2012you} \\
iCat                         & Cat     & 6  & \citep{bartneck2004your,heerink2010assessing,leite2008emotional,looije2010persuasive} \\
&&&\citep{shahid2014child,pereira2008icat,leyzberg2012physical}\\
Keepon                       & Chick   & 2  &  \citep{leyzberg2012physical} \\
Sony Aibo                    & Dog     & 4  & \citep{jung2004effects,lee2006physically} \\
Aesop Robot                  & Human   & 8  &  \citep{costa2016emotional}\\
Aldebaran NAO                & Human   & 6  &  \citep{fridin2014embodied,bremner2015speech,jost2014robot,kennedy2015robot}\\
&&&\citep{krogsager2014backchannel,ligthart2015selecting,looije2012help}\\
Bandit                       & Human   & 5  &  \citep{fasola2013socially,tapus2009role}\\
Darwin-OP                    & Human   & 6  &  \citep{brooks2012simulation}\\
eMuu                         & Human   & 1  &  \citep{bartneck2003interacting}\\
Honda ASIMO                  & Human   & 4  &  \citep{takeuchi2006comparison}\\
iCub                         & Human   & 7  &  \citep{lohan2010does,fischer2012levels} \\
KASPAR                       & Human   & 4  &  \citep{kose2009effects}\\
Kondo Kagaku         & Human   & 3  &  \citep{hasegawa2010role}\\
MIT Robot Head               & Human   & 5  &  \citep{kidd2004effect}\\
MIT AIDA                     & Human   & 2  &  \citep{williams2013reducing} \\
Nico Robot                   & Human   & 3  &  \citep{bainbridge2011benefits}\\
Nursebot                     & Human   & 4  &  \citep{kiesler2008anthropomorphic,powers2007comparing}\\
PaPeRo                       & Human   & 2  &  \citep{komatsu2010comparison}\\
Robata                       & Human   & 6  &  \citep{robins2006does}\\
Robothespian                 & Human   & 7  &  \citep{pan2016comparison} \\
Robotis Bioloid              & Human   & 5  &  \citep{jost2012robot}\\
Robovie-X                    & Human   & 7  &  \citep{nomura2009investigation}\\
Samsung April                & Human   & 6  &  \citep{jung2004effects,lee2006physically}\\
Stanford Kiosk Robot         & Kiosk   & 10 &  \citep{ju2010animate}\\
Robulab                      & Penguin & 7  &  \citep{wrobel2013effect} \\
Nabaztag                     & Rabbit  & 3  &  \citep{hoffmann2013investigating,zlotowski2010comparison}\\
NTT Lab Robot & Rabbit  & 5  &  \citep{shinozawa2003robots,shinozawa2007effect}\\
\hline
\end{tabular}
\end{table*}

\subsection{End-to-End Design of Socially Interactive Robots}
Using our characterization of the design space for socially interactive robots and the meta-results of the reviewed studies, we believe that the approach we present above can be effective in advising the design of new robots or selection of existing platforms for desired applications. The reviewed studies demonstrated that, for a given task, there are likely multiple social roles that robots can take. Similarly, the reviewed experiments demonstrated that the same social roles can be effectively implemented with different combinations of design metaphors and levels of abstraction.

Because ``optimality'' of socially interactive robot design is not a precise, quantitative process, the discussed approach can be applied to design or select not only robot embodiments for singular tasks but also for sets of tasks: first finding interaction strategies that have been shown to be effective for an individual tasks and then mapping those strategies to social roles and selecting the social role that is most effective for most (if not all) of the desired tasks \citep{kalegina2018characterizing,deng2018}. That social role can then be implemented and evaluated with a variety of design metaphors and levels of abstraction within the set of selected tasks.

\begin{figure}
  \centering
    \includegraphics[width=\textwidth]{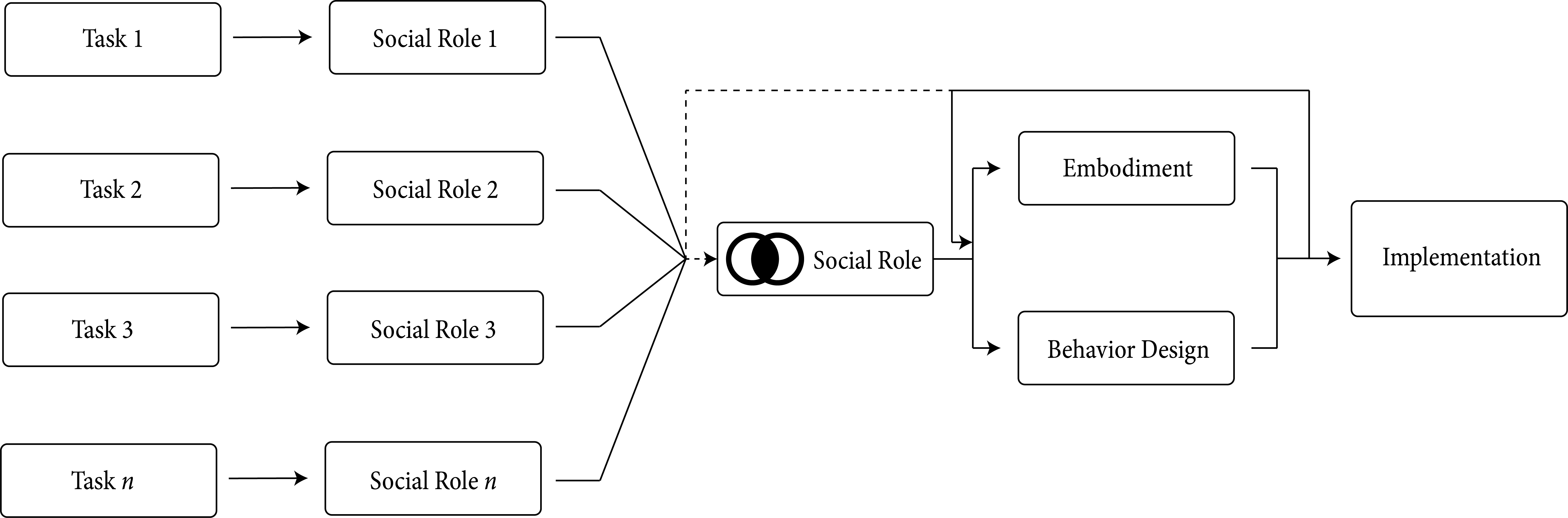}
  \caption{A characterization of the process for designing or selecting socially interactive robots for multiple tasks.}
  \label{fig:pipeline-complex}
\end{figure}

\subsection{Robot Embodiment Design in Practice}
Our formalization of the design and task spaces for socially interactive robots allows us to discuss and explore how robots are designed and used. We proposed a design process that defines the order in which design features should be decided and iterated on. Finally, using data from past research studies, we proposed two ways of visualizing and leveraging experimental data to drive future design decisions--specifically considering (1) \textit{the relationship between social roles of artificial agents used for different types of tasks} and (2) \textit{the mapping of differing levels of abstraction to social roles within each design metaphor}. To demonstrate how these steps come together in the design of new robots or the selection of an existing robot for a new task, we provide an example. While in the examples we do not iterate on any of the design decisions, as that requires evaluation and testing, in Figure \ref{fig:pipeline} we show where such iteration would take place on different design dimensions.

\subsubsection{Example: Grocer Store Robot}
Consider the example problem of needing to \textbf{design a robot that helps people decide what to purchase in a grocery store.} The robot needs to help people weigh the benefits of different food options against costs of those items. There is no objectively ``best'' combination of foods because features of foods can be more or less important to different people and some people may be more sensitive to price than others. Given that problem statement, this problem falls under the \textit{decision-making} task category.

\begin{figure}
  \centering
    \includegraphics[width=\textwidth]{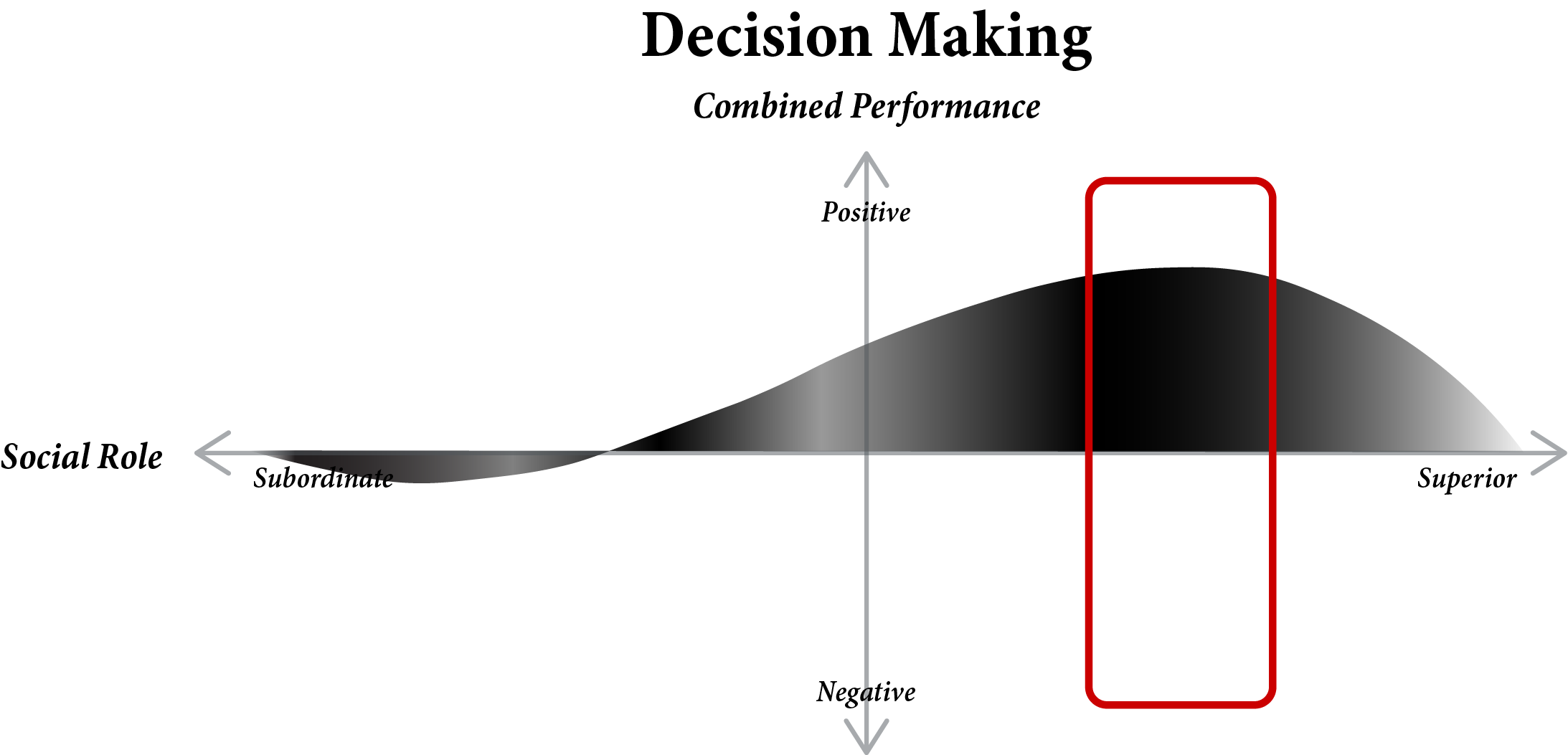}
  \caption{A visualization of the results from experiments with decision-making tasks plotted over the social role taken by the artificial agents and the combined performance within those experiments.}
  \label{fig:decision-mod}
\end{figure}

Taking into account the results from the surveyed experiments, we aim to make an educated guess as to the social role(s) to explore for the grocery store robot. Figure \ref{fig:decision-mod} shows the distribution of the decision-making task experiments plotted over \textit{performance} and \textit{social role} (with density of experiments shown by the overlaid gradient). The experimental results show, with relatively high confidence, that an agent with a social role somewhere between \textit{peer} and \textit{superior} seems to be the best choice. Given that, we then proceed to implement that social role into a robot embodiment, per Figure \ref{fig:pipeline}.

\begin{figure}
  \centering
    \includegraphics[width=\textwidth]{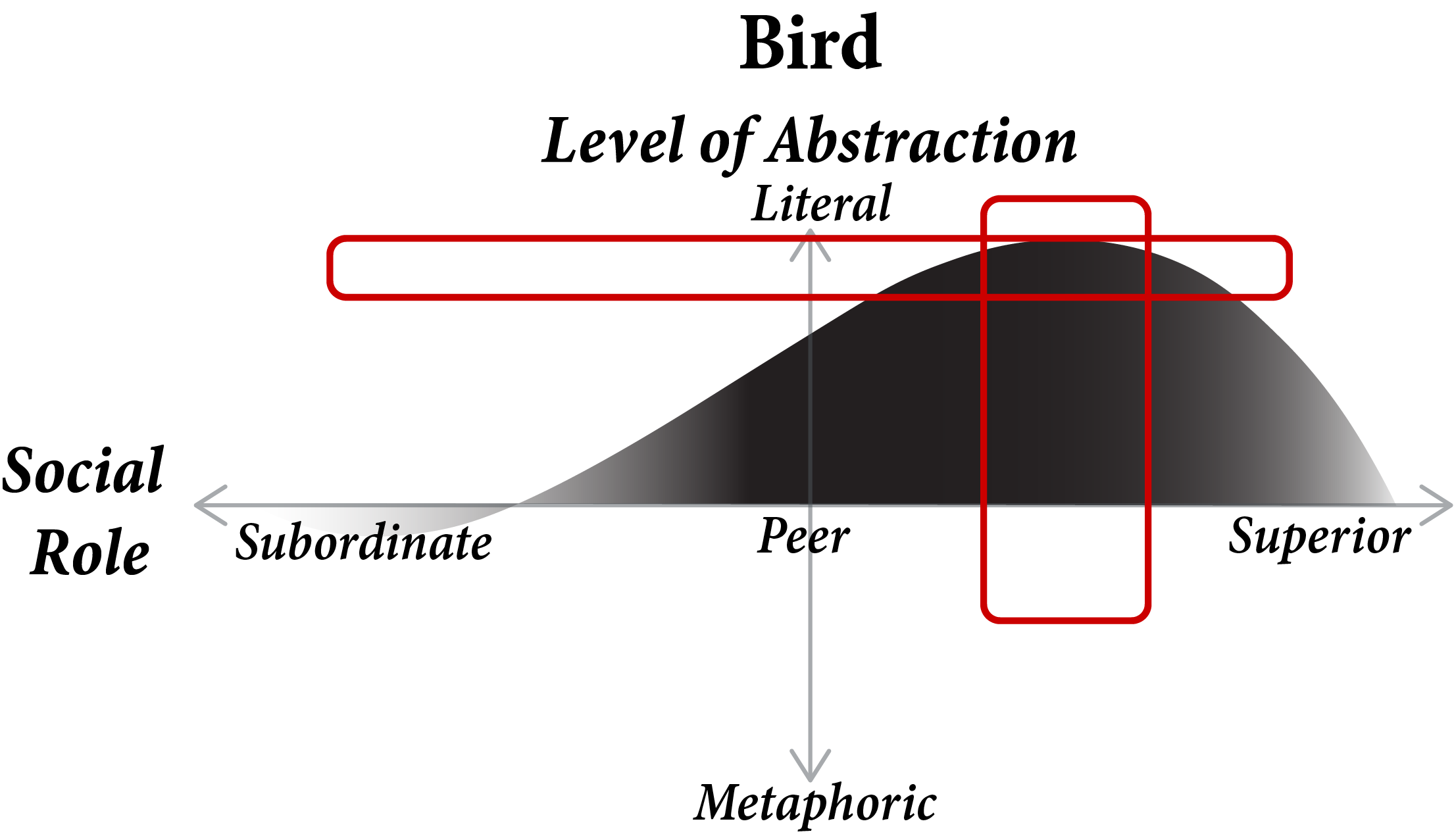}
  \caption{A visualization of the results from experiments with decision-making tasks plotted over the social role taken by the artificial agents and the combined performance within those experiments.}
  \label{fig:bird-mod}
\end{figure}

Because a social role can be implemented with different combinations of design metaphor and level of abstraction, we consider past effective design choices. The gradient in Figure \ref{fig:embodiment-selection} shows that \textit{birds}, \textit{humans}, \textit{cats} and \textit{cars} have been most frequently used for the peer-superior role. For this thought experiment, we arbitrarily select the \textit{bird} design metaphor to start with although in real applications it is best to iterate on design metaphors and select one based on other external constraints such as physical requirements for the robot.

The bird metaphor, like all other design metaphors, can be used to implement any social role with varying levels of efficacy. Based on reviewed robots and their usage, we visualize how the bird metaphor has been previously used (Figure \ref{fig:bird-mod}) and find that previous experiments have used a \textit{literal} instantiation of the bird design metaphor to implement a peer-superior role. Thus we select an existing robot fitting that metaphor and abstraction combination (from a database such as Table \ref{table:meta-robots}) \textit{or} design a new robot to fit that combination. The behaviors of the robot should then map to the appropriate levels of abstraction by changing how much it differs from the behaviors of the organic version of the metaphor.

Because the current dataset of embodiment-related experiments in the field of socially interactive robots is limited, we use all the data from studies covered in the review. The design goal can be more specific than a task--there may also be target user populations (e.g., children or elderly adults) or target demographic populations (e.g., different countries or cultures). Filtering results by those additional qualifiers (Table \ref{table:population}) allows for further tailoring the data to specific problem and context.

The characterization of the design space for socially interactive robots introduced in this work aims to facilitate a concrete discussion of past work, inform the selection and development of new robots for different applications, advise future experimental design, inspire novel applications, and help to improve the iterative design process of future robots, as the field of robotics continues to expand into new avenues of use.

\appendix

\chapter{Reviewed Studies}
\label{sec:appendix-studies}

%TA.1
\begin{table*}[h]
\centering
\caption{Reviewed studies labeled with the task category and social role of artificial agents used. Papers with multiple experiments are labeled with Exp. 1, 2, etc. and social role labeled with numeric scale of 1 (subordinate) to 9 (superior).\label{table:meta-overview}}
\Tiny
\begin{tabular}{lccc}
{\textbf Author (Year)} & {\textbf Task Category} & {\textbf Social Role} & {\textbf Reference}\\
\hline
Bainbridge et al. (2011), Exp. 1 & Performances/Actions & Superior/Peer (9) & \citep{bainbridge2011benefits} \\
\hline
Bainbridge et al. (2011), Exp. 2 & Performances/Actions & Superior/Peer (9) & \citep{bainbridge2011benefits} \\
\hline
Bainbridge et al. (2011), Exp. 3 & Performances/Actions & Superior/Peer (9) & \citep{bainbridge2011benefits} \\
\hline
Bartneck (2003) & Contests/Competition & Peer (5) & \citep{bartneck2003interacting} \\
\hline
Bartneck et al. (2004) & Intellective & Subordinate/Peer (3) & \citep{bartneck2004your} \\
\hline
Bremner and Leonards (2015) & Decision-Making & Subordinate/Peer (3) & \citep{bremner2015speech} \\
\hline
Brooks et al. (2012) & Performances/Actions & Superior/Peer (8) & \citep{brooks2012simulation} \\
\hline
Costa (2016), Exp. 1 & Creative & Superior/Peer (8) & \citep{costa2016emotional} \\
\hline
Costa (2016), Exp. 2 & Creative & Superior/Peer (8) & \citep{costa2016emotional} \\
\hline
Donahue and Scheutz (2015) & Performances/Actions & Subordinate (1) & \citep{donahue2015investigating} \\
\hline
Fasola and Mataric (2013) & Performances/Actions & Superior/Peer (8) & \citep{fasola2013socially} \\
\hline
Fischer et al. (2012), Exp. 1 & Intellective & Subordinate/Peer (2) & \citep{fischer2012levels} \\
\hline
Fischer et al. (2012), Exp. 2 & Intellective & Subordinate/Peer (2) & \citep{fischer2012levels} \\
\hline
Fischer et al. (2012), Exp. 3 & Intellective & Subordinate/Peer (2) & \citep{fischer2012levels} \\
\hline
Fridin and Belokopytov (2014) & Intellective & Superior/Peer (7) & \citep{fridin2014embodied} \\
\hline
Hasegawa et al. (2010) & Intellective & Superior/Peer (8) & \citep{hasegawa2010role} \\
\hline
Heerink et al. (2010) & Performances/Actions & Peer/Subordinate (2) & \citep{heerink2010assessing} \\
\hline
Hoffmann and Kr{\"a}mer (2013), Exp. 1 & Creative & Peer (5) & \citep{hoffmann2013investigating} \\
\hline
Hoffmann and Kr{\"a}mer (2013), Exp. 2 & Intellective & Peer/Superior (6) & \citep{hoffmann2013investigating} \\
\hline
Jost et al. (2014) & Intellective & Peer/Superior (6) & \citep{jost2014robot} \\
\hline
Jost et al. (2012), Exp. 1 & Contests/Competition & Peer (5) & \citep{jost2012ethological} \\
\hline
Jost et al. (2012), Exp. 2 & Contests/Competition & Peer (5) & \citep{jost2012ethological} \\
\hline
Ju and Sirkin (2010), Exp. 1 & Performances/Actions & Subordinate (1) & \citep{ju2010animate} \\
\hline
Ju and Sirkin (2010), Exp. 2 & Performances/Actions & Subordinate (1) & \citep{ju2010animate} \\
\hline
Jung and Lee (2004), Exp. 1 & Creative & Peer/Subordinate (2) & \citep{jung2004effects} \\
\hline
Jung and Lee (2004), Exp. 2 & Creative & Peer/Subordinate (2) & \citep{jung2004effects} \\
\hline

\end{tabular}
\end{table*}

\setcounter{table}{0}
\begin{table*}[t]
\centering
\caption{Continued}
\Tiny
\begin{tabular}{lccc}
{\textbf Author (Year)} & {\textbf Task Category} & {\textbf Social Role} & {\textbf Reference}\\
\hline
Kennedy et al. (2015) & Intellective & Superior/Peer (8) & \citep{kennedy2015robot}\\
\hline
Kidd and Breazeal (2004), Exp. 1 & Performances/Actions & Superior/Peer (8) & \citep{kidd2004effect} \\
\hline
Kidd and Breazeal (2004), Exp. 2 & Creative & Superior/Peer (7) & \citep{kidd2004effect} \\
\hline
Kiesler et al. (2008) & Decision Making & Peer (5) & \citep{kiesler2008anthropomorphic} \\
\hline
Komatsu et al. (2010), Exp. 1 & Intellective & Peer/Subordinate (3) & \citep{komatsu2010comparison} \\
\hline
Komatsu et al. (2010), Exp. 2 & Intellective & Peer/Subordinate (3) & \citep{komatsu2010comparison} \\
\hline
Kose et al. (2009) & Performances/Actions & Peer (5) & \citep{kose2009effects} \\
\hline
Krogsager et al. (2014) & Creative & Peer (5) & \citep{krogsager2014backchannel} \\
\hline
Lee et al. (2006), Exp. 1 & Creative & Peer (5) & \citep{lee2006physically} \\
\hline
Lee et al. (2006), Exp. 2 & Creative & Peer (5) & \citep{lee2006physically} \\
\hline
Lee et al. (2015) & Decision Making & Peer/Subordinate (4) & \citep{lee2015impact} \\
\hline
Leite et al. (2008) & Cognitive Conflict & Peer (5) & \citep{leite2008emotional} \\
\hline
Levy-Tzedek et al. (2017) & Performances/Actions & Superior/Peer (7) & \citep{levy2017robotic} \\
\hline
Leyzberg et al. (2012) & Contests/Competition & Peer/Superior (7) & \citep{leyzberg2012physical} \\
\hline
Li and Chignell (2011) & Decision Making & Subordinate (1)& \citep{li2011communication} \\
\hline
Ligthart and Truong (2015) & Cognitive Conflict & Peer/Superior (6) & \citep{ligthart2015selecting} \\
\hline
Lohan et al. (2010) & Intellective & Peer/Subordinate (4)& \citep{lohan2010does} \\
\hline
Looije et al. (2010) & Cognitive Conflict & Peer/Superior (7) & \citep{looije2010persuasive} \\
\hline
Looije et al. (2012) & Contests/Competition & Peer/Superior (7) & \citep{looije2012help} \\
\hline
Nomura (2009) & Performances/Actions & Superior/Peer (8) & \citep{nomura2009investigation} \\
\hline
Pan and Steed (2016) & Cognitive Conflict & Superior/Peer (8) & \citep{pan2016comparison} \\
\hline
Pereira et al. (2008) & Cognitive Conflict & Peer (5)& \citep{pereira2008icat} \\
\hline
Powers et al. (2007) & Mixed Motive & Peer/Superior (7)& \citep{powers2007comparing} \\
\hline
Robins et al. (2006) & Mixed Motive & Peer/Subordinate (4)& \citep{robins2006does} \\
\hline
Segura et al. (2012) & Performances/Actions & Superior/Peer (8)& \citep{segura2012you} \\
\hline
Shahid et al. (2014) & Cognitive Conflict & Peer (5) & \citep{shahid2014child} \\
\hline
Shinozawa and Reeves (2003), Exp. 1 & Mixed Motive & Peer/Subordinate (4) & \citep{shinozawa2003robots} \\
\hline
Shinozawa and Reeves (2003), Exp. 2 & Planning & Superior/Peer (6) & \citep{shinozawa2003robots} \\
\hline
Shinozawa and Reeves (2003), Exp. 3 & Intellective & Peer (5) & \citep{shinozawa2003robots} \\
\hline
Shinozawa et al. (2007) & Decision Making & Peer/Superior (7) & \citep{shinozawa2007effect} \\
\hline
Short et al. (2017) & Creative & Subordinate/Peer (4) & \citep{Short-2017-998} \\
\hline
Takeuchi et al. (2006) & Intellective & Superior/Peer (8) & \citep{takeuchi2006comparison} \\
\hline
Tapus et al. (2009) & Mixed Motive & Peer/Superior (7) & \citep{tapus2009role} \\
\hline
Vossen et al. (2009) & Mixed Motive & Peer/Superior (7) & \citep{tapus2009role} \\
\hline
Wainer et al. (2006) & Intellective & Superior (9) & \citep{wainer2006role} \\
\hline
Wainer et al. (2007) & Intellective & Superior (9) & \citep{wainer2007embodiment} \\
\hline
Williams et al. (2013) & Planning & Superior (9) & \citep{williams2013reducing} \\
\hline
Wrobel et al. (2013) & Contests/Competition & Peer (5) & \citep{wrobel2013effect} \\
\hline
Zlotowski (2010) & Intellective & Peer/Superior (7) & \citep{zlotowski2010comparison} \\
\hline
\end{tabular}
\end{table*}

\begin{table*}
\centering
\caption{The physical robot platforms and their virtual counterparts used in the reviewed studies.\label{table:platforms}}
\Tiny
\begin{tabular}{lccc}
{\textbf Author (Year)} & {\textbf Physical Agent} & {\textbf Virtual Agent}\\ \hline
Bainbridge et al. (2011), Exp. 1    & Nico                 & Nico                           \\ \hline
Bainbridge et al. (2011), Exp. 2    & Nico                 & Live Video of Nico             \\ \hline
Bainbridge et al. (2011), Exp. 3    & Nico                 & Live Video of Nico             \\\hline
Bartneck (2003)                     & eMuu                 & Virtual eMuu                   \\\hline
Bartneck et al. (2004)              & iCat                 & Virtual iCat                   \\\hline
Bremner and Leonard (2015)          & NAO                  & Live Video of Human            \\\hline
Brooks et al, (2012)                & Darwin-OP            & Manoi Animation                \\\hline
Costa (2014), Exp. 1                & Aesop Robot          & Greta Animation                \\\hline
Costa (2014), Exp. 2                & Aesop Robot          & Greta Animation                \\ \hline
Donahue and Scheutz (2015)          & Pioneer 2DX          & Virtual Pioneer 2DX            \\\hline
Fasola \& Mataric (2013)            & Bandit               & Virtual Bandit                 \\\hline
Fischer et al. (2012), Exp. 1       & iCub II              & Akachan                        \\\hline
Fischer et al. (2012), Exp. 2       & iCub II              & Akachan                        \\\hline
Fischer et al. (2012), Exp. 3       & iCub II              & Akachan                        \\\hline
Fridin and Belokopytov (2014)       & NAO                  & Virtual NAO                    \\\hline
Hasegawa et al. (2010)              & Kondo Kagaku KHR2-HV & NUMACK                         \\\hline
Heerink et al. (2009)               & iCat                 & IIE Annie                      \\\hline
Hoffmann \& Kr{\"a}mer (2013), Exp. 1   & Nabaztag             & Virtual Nabaztag               \\\hline
Hoffmann \& Kr{\"a}mer (2013), Exp. 2   & Nabaztag             & Virtual Nabaztag               \\\hline
Jost el al. (2014)                  & NAO         &                                \\\hline
Jost et al. (2012), Exp. 1          & Robotis Bioloid      & Telecom GRETA                  \\\hline
Jost et al. (2012), Exp. 2          & Robotis Bioloid      & Telecom GRETA                  \\\hline
Ju and Sirkin (2010), Exp. 1        & Kiosk Robot with Arm & Kiosk Robot with Projected Arm \\\hline
Ju and Sirkin (2010), Exp. 2        & Kiosk Robot with Arm & Kiosk Arm with on-screen Arm   \\\hline
Jung and Lee (2004), Exp. 1         & Sony Aibo            & Virtual Sony Aibo              \\\hline
Jung and Lee (2004), Exp. 2         & Samsung April        & Virtual Samsung April          \\\hline
Kennedy et al. (2015)               & NAO                  & Virtual NAO                    \\\hline
Kidd \& Breazeal (2004), Exp. 1     & Robot Eyes           & Virtual Eyes                   \\\hline
Kidd \& Breazeal (2004), Exp. 2     & MIT Robot Head       & Virtual MIT Robot Head         \\\hline
Kiesler et al. (2008)               & Nursebot             & Virtual Nursebot               \\\hline
Komatsu et al. (2010), Exp. 1       & PaPeRo               & RobotStudio                    \\\hline
Komatsu et al. (2010), Exp. 2       & PaPeRo               & RobotStudio                    \\\hline
Kose et al. (2009)                  & KASPAR               & Virtual KASPAR                 \\\hline
Krogsager et al. (2014)             & Aldebaran NAO        & Telepresent NAO                \\\hline
Lee et al. (2006), Exp. 1           & Sony Aibo            & Virtual Aibo                   \\\hline
Lee et al. (2006), Exp. 2           & April                & Virtual April                  \\\hline
Lee et al. (2015)                   & Humanoid Robot       & Virtual Humanoid Robot         \\\hline
Leite et al. (2008)					& iCat				   & Virtual iCat 					\\\hline
Levy-Tzedek et al. (2017)           & Kinova Arm           & Video of Kinova Arm            \\\hline
Leyzberg et al. (2012)              & Keepon               & Video of Keepon                \\\hline
Li and Chignell (2011)              & Keio U Robotphone    & Virtual Keio U Robotphone      \\\hline
Ligthart and Truong (2015)          & NAO                  & Virtual NAO                    \\\hline
Lohan et al. (2010)                 & iCub                 & Akachan                        \\\hline
Looije et al. (2012)                & NAO                  & Virtual NAO                    \\\hline
Looije, Neerincx, \& Cnossen (2010) & iCat                 & Virtual iCat                   \\\hline
Nomura (2009)                       & Robovie-X            & Virtual Robovie-X              \\\hline
Pan and Steed (2016)                & Robothespian         & Virtual Human Character        \\\hline
Pereira et al. (2008)               & iCat robot           & Virtual iCat                   \\\hline
Powers et al. (2007)                & Nursebot             & Projected Virtual Agent        \\\hline
Robins et al. (2006)                & Robata               & Passive Robata Doll            \\\hline
Segura et al. (2012)                & Pioneer P3AT         & Virtual P3AT Head              \\\hline
\end{tabular}
\end{table*}

\setcounter{table}{1}
\begin{table*}
\centering
\caption{Continued}
\Tiny
\begin{tabular}{lccc}
{\textbf Author (Year)} & {\textbf Physical Agent} & {\textbf Virtual Agent}\\ \hline
Shahid et al. (2014)                & iCat                 & Human                          \\\hline
Shinozawa and Reeves (2002), Exp. 1 & NTT Lab Robot        & Video of Lab Robot             \\\hline
Shinozawa and Reeves (2002), Exp. 2 & NTT Lab Robot        & Video of Lab Robot             \\\hline
Shinozawa and Reeves (2002), Exp. 3 & NTT Lab Robot        & Video of Lab Robot             \\\hline
Shinozawa et al. (2007)				& NTT Lab Robot		   & Video of Lab Robot				\\\hline
Short et al. (2017)                 & Bandit on Pioneer    & Pioneer with Bubble machine    \\\hline
Takeuchi et al. (2006)              & Honda ASIMO          & Microsoft Peedy                \\\hline
Tapus, Tapus \& Mataric (2009)      & Bandit               & Virtual Bandit                 \\\hline
Vossen et al. (2009)                & iCat                 & Voice only                     \\\hline
Wainer et al. (2006)                & Pioneer 2DX          & Virtual Pioneer 2DX            \\\hline
Wainer et al. (2007)                & Pioneer 2DX          & Virtual Pioneer 2DX            \\\hline
Williams et al. (2013)              & MIT AIDA             & AIDA on-screen App             \\\hline
Wrobel et al. (2013)                & Robulab              & Virtual Greta                  \\\hline
Zlotowski (2010)                    & Nabaztag             & Virtual Nabaztag               \\\hline
\end{tabular}
\end{table*}

\begin{table*}
\centering
\caption{The results of the reviewed studies broken down by \textit{task performance differences} and \textit{interaction performance differences.}\label{table:results}}
\Tiny
\begin{tabular}{lccc}
{\textbf Author (Year)} & {\textbf Task Performance} & {\textbf Interaction Performance}\\ \hline
Bainbridge et al. (2011), Exp. 1    & PE \textgreater VE  & PE \textgreater VE  \\
Bainbridge et al. (2011), Exp. 2    & PE \textgreater VE  & PE \textgreater VE  \\
Bainbridge et al. (2011), Exp. 3    & PE \textgreater VE  & PE \textgreater VE  \\
Bartneck (2003)                     & PE \textgreater VE  & PE = VE             \\
Bartneck et al. (2004)              & PE = VE             & PE = VE             \\
Bremner and Leonard (2015)          & PE = VE             & N/A                 \\
Brooks et al, (2012)                & PE  \textgreater VE & N/A                 \\
Costa (2014), Exp. 1                & PE \textgreater VE  & PE = VE             \\
Costa (2014), Exp. 2                & PE \textgreater VE  & PE = VE             \\
Donahue and Scheutz (2015)          & PE \textgreater VE  & N/A                 \\
Fasola \& Mataric (2013)            & PE = VE             & PE \textgreater VE  \\
Fischer et al. (2012), Exp. 1       & PE  \textgreater VE & N/A                 \\
Fischer et al. (2012), Exp. 2       & PE  \textgreater VE & N/A                 \\
Fischer et al. (2012), Exp. 3       & PE  \textgreater VE & N/A                 \\
Fridin and Belokopytov (2014)       & PE = VE             & PE \textgreater VE  \\
Hasegawa et al. (2010)              & PE = VE             & PE \textgreater VE  \\
Heerink et al. (2009)               & N/A                 & PE \textgreater VE  \\
Hoffmann \& Kr{\"a}mer (2013), Exp. 1   & PE = VE             & PE \textless VE     \\
Hoffmann \& Kr{\"a}mer (2013), Exp. 2   & PE = VE             & PE \textgreater VE  \\
Jost el al. (2014)                  & PE \textgreater VE  & N/A                 \\
Jost et al. (2012), Exp. 1          & PE \textgreater VE  & PE \textgreater VE  \\
Jost et al. (2012), Exp. 2          & PE \textgreater VE  & PE \textgreater VE  \\
Ju and Sirkin (2010), Exp. 1        & PE \textgreater VE  & PE \textless VE     \\
Ju and Sirkin (2010), Exp. 2        & PE \textgreater VE  & PE \textless VE     \\
Jung and Lee (2004), Exp. 1         & N/A                 & PE \textgreater VE  \\
Jung and Lee (2004), Exp. 2         & N/A                 & PE \textgreater VE  \\
Kennedy et al. (2015)               & PE \textgreater VE  & PE \textgreater VE  \\
Kidd \& Breazeal (2004), Exp. 1     & N/A                 & PE \textgreater VE  \\
Kidd \& Breazeal (2004), Exp. 2     & N/A                 & PE \textgreater VE  \\
Kiesler et al. (2008)               & PE \textgreater VE  & PE \textgreater VE  \\
Komatsu et al. (2010), Exp. 1       & PE = VE             & PE \textgreater VE  \\
Komatsu et al. (2010), Exp. 2       & PE = VE             & PE \textgreater VE  \\
Kose et al. (2009)                  & PE  \textgreater VE & PE = VE             \\
Krogsager et al. (2014)             & PE \textless VE     & PE \textless VE     \\
Lee et al. (2006), Exp. 1           & N/A                 & PE \textgreater VE  \\
Lee et al. (2006), Exp. 2           & N/A                 & PE \textless VE     \\
Lee et al. (2015)                   & N/A                 & PE = VE             \\
Leite et al. (2008)                 & N/A                 & PE \textgreater VE  \\
Levy-Tzedek et al. (2017)           & PE \textgreater VE  & PE \textgreater VE  \\
Leyzberg et al. (2012)              & PE \textgreater VE  & PE \textgreater VE  \\
Li and Chignell (2011)              & PE = VE             & N/A                 \\
Ligthart and Truong (2015)          & N/A                 & PE = VE             \\
Lohan et al. (2010)                 & N/A                 & PE \textgreater VE  \\
Looije et al. (2012)                & N/A                 & PE \textgreater VE  \\
Looije, Neerincx, \& Cnossen (2010) & PE \textgreater VE  & PE \textless VE     \\
Nomura (2009)                       & PE = VE             & PE = VE             \\
Pan and Steed (2016)                & PE \textgreater VE  & PE \textgreater VE  \\
Pereira et al. (2008)               & N/A                 & PE \textgreater VE  \\
Powers et al. (2007)                & PE \textless VE     & PE \textgreater VE  \\
Robins et al. (2006)                & PE \textgreater VE  & PE \textgreater VE  \\
Segura et al. (2012)                & PE \textgreater VE  & PE \textgreater VE  \\
\hline
\end{tabular}
\end{table*}

\makeatletter
\setlength{\@fptop}{0pt}
\makeatother

\setcounter{table}{2}
\begin{table*}
\centering
\caption{Continued}
\Tiny
\begin{tabular}{lccc}
{\textbf Author (Year)} & {\textbf Task Performance} & {\textbf Interaction Performance}\\ \hline
Shahid et al. (2014)                & PE \textgreater VE  & PE \textgreater VE  \\
Shinozawa and Reeves (2002), Exp. 1 & PE \textgreater VE  & PE \textgreater VE  \\
Shinozawa and Reeves (2002), Exp. 2 & PE \textgreater VE  & PE \textgreater VE  \\
Shinozawa and Reeves (2002), Exp. 3 & PE \textgreater VE  & PE \textgreater VE  \\
Shinozawa et al. (2007)             & PE  \textgreater VE & PE  \textgreater VE \\
Short et al. (2017)                 & PE  \textgreater VE & PE \textgreater VE  \\
Takeuchi et al. (2006)              & PE \textgreater VE  & PE \textgreater VE  \\
Tapus, Tapus \& Mataric (2009)      & PE \textgreater VE  & PE \textgreater VE  \\
Vossen et al. (2009)                & PE \textgreater VE  & PE \textgreater VE  \\
Wainer et al. (2006)                & PE \textgreater VE  & PE \textgreater VE  \\
Wainer et al. (2007)                & PE \textgreater VE  & PE \textgreater VE  \\
Williams et al. (2013)              & PE \textgreater VE  & PE \textgreater VE  \\
Wrobel et al. (2013)                & PE \textgreater VE  & PE \textgreater VE  \\
Zlotowski (2010)                    & PE \textless VE     & PE \textless VE     \\
\hline
\end{tabular}
\end{table*}

\begin{table*}
\centering
\caption{Demographic characteristics of the participant pools in the reviewed studies.\label{table:population}}
\Tiny
\begin{tabular}{lccc}
{\textbf Author (Year)} & {\textbf \textit{n}} & {\textbf Age Group} & {\textbf Country}\\ \hline
Bainbridge et al. (2011), Exp. 1    & 59  & Adults              & US                       \\
Bainbridge et al. (2011), Exp. 2    & 59  & Adults              & US                       \\
Bainbridge et al. (2011), Exp. 3    & 59  & Adults              & US                       \\
Bartneck (2003)                     & 53  & Adults              & Netherlands              \\
Bartneck et al. (2004)              & 56  & Adults              & Netherlands              \\
Bremner and Leonard (2015)          & 22  & Adults              & UK                       \\
Brooks et al, (2012)                & 11  & Adults              & US                       \\
Costa (2014), Exp. 1                & 20  & Adults              & United Arab Emirates     \\
Costa (2014), Exp. 2                & 40  & Adults              & United Arab Emirates     \\
Donahue and Scheutz (2015)          & 55  & Adults              & US                       \\
Fasola \& Mataric (2013)            & 33  & Elderly Adults      & US                       \\
Fischer et al. (2012), Exp. 1       & 38  & Adults              & Germany                  \\
Fischer et al. (2012), Exp. 2       & 14  & Adults              & Germany                  \\
Fischer et al. (2012), Exp. 3       & 36  & Adults              & Germany                  \\
Fridin and Belokopytov (2014)       & 13  & Children            & Israel                   \\
Hasegawa et al. (2010)              & 75  & Children            & Japan                    \\
Heerink et al. (2009)               & 40  & Elderly Adults      & Netherlands              \\
Hoffmann \& Kr{\"a}mer (2013), Exp. 1   & 83  & Adults              & Germany                  \\
Hoffmann \& Kr{\"a}mer (2013), Exp. 2   & 83  & Adults              & Germany                  \\
Jost el al. (2014)                  & 67  & Children and Adults & France                   \\
Jost et al. (2012), Exp. 1          & 51  & Children            & France                   \\
Jost et al. (2012), Exp. 2          & 52  & Children            & France                   \\
Ju and Sirkin (2010), Exp. 1        & 179 & Adults              & US                       \\
Ju and Sirkin (2010), Exp. 2        & 457 & Adults              & US                       \\
Jung and Lee (2004), Exp. 1         & 36  & Adults              & US                       \\
Jung and Lee (2004), Exp. 2         & 32  & Adults              & US                       \\
Kennedy et al. (2015)               & 28  & Children            & EU                       \\
Kidd \& Breazeal (2004), Exp. 1     & 32  & Adults              & US                       \\
Kidd \& Breazeal (2004), Exp. 2     & 82  & Adults              & US                       \\
Kiesler et al. (2008)               & 113 & Adults              & US                       \\
Komatsu et al. (2010), Exp. 1       & 20  & Children            & Japan                    \\
Komatsu et al. (2010), Exp. 2       & 40  & Children            & Japan                    \\
Kose et al. (2009)                  & 66  & Children            & UK                       \\
Krogsager et al. (2014)             & 9   & Children            & Denmark                  \\
Lee et al. (2006), Exp. 1           & 32  & Adults              & US                       \\
Lee et al. (2006), Exp. 2           & 32  & Adults              & US                       \\
Lee et al. (2015)                   & 24  & Adults              & US                       \\
Leite et al. (2008)                 & 9   & Children and Adults & Portugal                 \\
Levy-Tzedek et al. (2017)           & 22  & Adults              & Israel                   \\
Leyzberg et al. (2012)              & 100 & Adults              & US                       \\
Li and Chignell (2011)              & 16  & Adults              & Japan                    \\
Ligthart and Truong (2015)          & 40  & Adults              & Netherlands              \\
Lohan et al. (2010)                 & 28  & Adults              & Germany                  \\
Looije et al. (2012)                & 11  & Children            & Netherlands              \\
Looije, Neerincx, \& Cnossen (2010) & 24  & Adults              & Netherlands              \\
Nomura (2009)                       & 37  & Adults              & Japan                    \\
Pan and Steed (2016)                & 24  & Adults              & UK                       \\
Pereira et al. (2008)               & 18  & Children            & Portugal                 \\
Powers et al. (2007)                & 113 & Adults              & US                       \\
Robins et al. (2006)                & 4   & Children            & UK                       \\
Segura et al. (2012)                & 42  & Adults              & UK                       \\
\hline
\end{tabular}
\end{table*}

\setcounter{table}{3}
\begin{table*}
\centering
\caption{Continued}
\Tiny
\begin{tabular}{lccc}
{\textbf Author (Year)} & {\textbf \textit{n}} & {\textbf Age Group} & {\textbf Country}\\ \hline
Shahid et al. (2014)                & 112 & Children            & Netherlands and Pakistan \\
Shinozawa and Reeves (2002), Exp. 1 & 72  & Adults              & Japan and US             \\
Shinozawa and Reeves (2002), Exp. 2 & 72  & Adults              & Japan and US             \\
Shinozawa and Reeves (2002), Exp. 3 & 72  & Adults              & Japan and US             \\
Shinozawa et al. (2007)             & 178 & Adults              & Japan                    \\
Short et al. (2017)                 & 6   & Children            & US                       \\
Takeuchi et al. (2006)              & 31  & Adults              & Japan                    \\
Tapus, Tapus \& Mataric (2009)      & 3   & Elderly Adults      & US                       \\
Vossen et al. (2009)                & 76  & Adults              & Netherlands              \\
Wainer et al. (2006)                & 11  & Adults              & US                       \\
Wainer et al. (2007)                & 21  & Adults              & US                       \\
Williams et al. (2013)              & 44  & Adults              & US                       \\
Wrobel et al. (2013)                & 19  & Adults              & France                   \\
Zlotowski (2010)                    & 16  & Adults              & Finland                  \\
\hline
\end{tabular}
\end{table*}

\begin{table*}
\centering
\caption{Reviewed studies labeled with observed measures used (task performance, interaction performance, individual behavior) and whether or not self-reported measures were implemented.\label{table:measures}}
\Tiny
\begin{tabular}{lccc}
&& {\textbf Self-Reported }\\
{\textbf Author (Year)} & {\textbf Observed Measures} & {\textbf Measures}\\ \hline
Bainbridge et al. (2011), Exp 1     & Task Performance, Interaction Performance   & Yes \\
Bainbridge et al. (2011), Exp. 2    & Task Performance, Interaction Performance   & Yes \\
Bainbridge et al. (2011), Exp. 3    & Task Performance, Interaction Performance   & Yes \\
Bartneck (2003)                     & Task Performance                            & Yes \\
Bartneck et al. (2004)              & Task Performance                            & Yes \\
Bremner and Leonard (2015)          & Individual Behavior                         & No  \\
Brooks et al, (2012)                & Task Performance, Interaction Performance   & No  \\
Costa (2014), Exp. 1                & Individual Behavior, Interaction Performance & Yes \\
Costa (2014), Exp. 2                & individual Behavior, Interaction Performance & Yes \\
Donahue and Scheutz (2015)          & Individual Behavior                         & No  \\
Fasola \& Mataric (2013)            & Task Performance                            & Yes \\
Fischer et al. (2012), Exp. 1       & Task Performance, Interaction Performance   & No  \\
Fischer et al. (2012), Exp. 2       & Task Performance, Interaction Performance   & No  \\
Fischer et al. (2012), Exp. 3       & Task Performance, Interaction Performance   & No  \\
Fridin and Belokopytov (2014)       & Task Performance, Interaction Performance   & Yes \\
Hasegawa et al. (2010)              & Task Performance                            & Yes \\
Heerink et al. (2009)               & Individual Behavior                         & Yes \\
Hoffmann \& Kr{\"a}mer (2013), Exp. 1   & Task Performance                            & Yes \\
Hoffmann \& Kr{\"a}mer (2013), Exp. 2   & Task Performance                            & Yes \\
Jost el al. (2014)                  & Individual Behavior, Interaction Performance & No  \\
Jost et al. (2012), Exp. 1          & Task Performance, Interaction Performance   & Yes \\
Jost et al. (2012), Exp. 2          & N/A                                         & Yes \\
Ju and Sirkin (2010), Exp. 1        & Task Performance                            & Yes \\
Ju and Sirkin (2010), Exp. 2        & Task Performance                            & Yes \\
Jung and Lee (2004), Exp. 1         & N/A                                         & Yes \\
Jung and Lee (2004), Exp. 2         & N/A                                         & Yes \\
Kennedy et al. (2015)               & Task Performance                            & Yes \\
Kidd \& Breazeal (2004), Exp. 1     & Individual Behavior, Interaction Performance & Yes \\
Kidd \& Breazeal (2004), Exp. 2     & N/A                                         & Yes \\
Kiesler et al. (2008)               & Task Performance                            & Yes \\
Komatsu et al. (2010), Exp. 1       & Task Performance, Interaction Performance   & No  \\
Komatsu et al. (2010), Exp. 2       & Task Performance, Interaction Performance   & No  \\
Kose et al. (2009)                  & Task Performance, Interaction Performance   & Yes \\
Krogsager et al. (2014)             & Task Performance                            & Yes \\
Lee et al. (2006), Exp. 1           & N/A                                         & Yes \\
Lee et al. (2006), Exp. 2           & N/A                                         & Yes \\
Lee et al. (2015)                   & N/A                                         & Yes \\
Leite et al. (2008)                 & N/A                                         & Yes \\
Levy-Tzedek et al. (2017)           & Task Performance                            & Yes \\
Leyzberg et al. (2012)              & Task Performance                            & Yes \\
Li and Chignell (2011)              & Task Performance                            & Yes \\
Ligthart and Truong (2015)          & N/A                                         & Yes \\
Lohan et al. (2010)                 & Task Performance, Interaction Performance   & No  \\
Looije et al. (2012)                & Task Performance, Interaction Performance   & Yes \\
Looije, Neerincx, \& Cnossen (2010) & N/A                                         & Yes \\
Nomura (2009)                       & Task Performance                            & Yes \\
Pan and Steed (2016)                & Task Performance, Interaction Performance   & Yes \\
Pereira et al. (2008)               & N/A                                         & Yes \\
Powers et al. (2007)                & Task Performance, Interaction Performance   & Yes \\
Robins et al. (2006)                & Task Performance, Interaction Performance   & Yes \\
Segura et al. (2012)                & Task Performance, Interaction Performance   & Yes \\
\hline
\end{tabular}
\end{table*}

\setcounter{table}{4}
\begin{table*}
\centering
\caption{Continued}
\Tiny
\begin{tabular}{lccc}
&& {\textbf Self-Reported }\\
{\textbf Author (Year)} & {\textbf Observed Measures} & {\textbf Measures}\\ \hline
Shahid et al. (2014)                & Task Performance, Interaction Performance   & Yes \\
Shinozawa and Reeves (2002), Exp. 1 & Task Performance                            & Yes \\
Shinozawa and Reeves (2002), Exp. 2 & Task Performance                            & Yes \\
Shinozawa and Reeves (2002), Exp. 3 & Task Performance                            & Yes \\
Shinozawa et al. (2007)             & Task Performance                            & Yes \\
Short et al. (2017)                 & Task Performance, Interaction Performance   & Yes \\
Takeuchi et al. (2006)              & N/A                                         & Yes \\
Tapus, Tapus \& Mataric (2009)      & Task Performance, Interaction Performance   & Yes \\
Vossen et al. (2009)                & Task Performance                            & Yes \\
Wainer et al. (2006)                & Task Performance                            & Yes \\
Wainer et al. (2007)                & Task Performance                            & Yes \\
Williams et al. (2013)              & Task Performance, Interaction Performance   & Yes \\
Wrobel et al. (2013)                & N/A                                         & Yes \\
Zlotowski (2010)                    & Task Performance                            & Yes \\
\hline
\end{tabular}
\end{table*}

\backmatter  % references

%KG\bibliographystyle{plainnat}
%KG\bibliography{sample}

\end{document}